\pdfoutput=1

\documentclass[11pt]{article}

\usepackage[preprint]{coling}

\usepackage{times}
\usepackage{latexsym}

\usepackage[T1]{fontenc}

\usepackage[utf8]{inputenc}

\usepackage{microtype}

\usepackage{inconsolata}

\usepackage{graphicx}

\usepackage{tablefootnote}

\usepackage{color, colortbl}
\definecolor{verylightgray}{rgb}{0.9,0.9,0.9}
\definecolor{gray}{rgb}{0.5,0.5,0.5}
\definecolor{pygreen}{rgb}{0.0, 0.5, 0.0}
\definecolor{pyred}{rgb}{0.7, 0.0, 0.0}
\definecolor{pyblue}{rgb}{0.0, 0.0, 0.7}
\definecolor{pygray}{rgb}{0.5, 0.5, 0.5}
\definecolor{pydarkgray}{rgb}{0.3, 0.3, 0.3}

\definecolor{color1}{HTML}{006EB8}
\definecolor{color2}{HTML}{009B55}

\usepackage{wrapfig}

\usepackage[symbol]{footmisc}

\usepackage[T1]{fontenc}
\usepackage{multirow}
\usepackage{microtype}
\usepackage{hyperref}
\hypersetup{colorlinks=true,urlcolor=color2,linkcolor=color2,citecolor=color2}
\usepackage{url}
\usepackage{booktabs}
\usepackage{graphicx}
\usepackage{subcaption}
\usepackage[whole]{bxcjkjatype}
\usepackage{amsmath} 
\usepackage[inline]{enumitem}

\newcommand{\system}{\textsc{Aurora-M}}

\title{\textsc{\system}: Open Source Continual Pre-training for Multilingual Language and Code}

\author{
Taishi Nakamura$^{*1}$, 
Mayank Mishra$^{*2}$, 
Simone Tedeschi$^{*3,4}$, 
Yekun Chai$^5$ \\
\textbf{Jason T Stillerman}, 
\textbf{Felix Friedrich}$^{6,7}$, 
\textbf{Prateek Yadav}$^8$, 
\textbf{Tanmay Laud}, \\
\textbf{Vu Minh Chien}$^9$, 
\textbf{Terry Yue Zhuo}$^{10,11}$, 
\textbf{Diganta Misra}$^{12,13}$, 
\textbf{Ben Bogin}$^{14}$, \\
\textbf{Xuan-Son Vu}$^{15,16,17}$, 
\textbf{Marzena Karpinska}$^{18}$, 
\textbf{Arnav Varma Dantuluri}, 
\textbf{Wojciech Kusa}$^{33}$, \\
\textbf{Tommaso Furlanello}, 
\textbf{Rio Yokota}$^1$, 
\textbf{Niklas Muennighoff}, 
\textbf{Suhas Pai}$^{19}$, \\
\textbf{Tosin Adewumi}$^{20}$, 
\textbf{Veronika Laippala}, 
\textbf{Xiaozhe Yao}$^{21}$, 
\textbf{Adalberto Junior}, \\
\textbf{Alpay Ariyak}$^{22,23}$, 
\textbf{Aleksandr Drozd}$^{24}$, 
\textbf{Jordan Clive}$^{25}$, 
\textbf{Kshitij Gupta}$^{12}$, \\
\textbf{Liangyu Chen}, 
\textbf{Qi Sun}$^1$, 
\textbf{Ken Tsui}, 
\textbf{Noah Persaud}, \\
\textbf{Nour Fahmy}, 
\textbf{Tianlong Chen}$^8$, 
\textbf{Mohit Bansal}$^8$, 
\textbf{Nicolò Monti}$^{26}$, \\
\textbf{Tai Dang}$^{18}$, 
\textbf{Ziyang Luo}$^{27}$, 
\textbf{Tien-Tung Bui}$^{28}$, 
\textbf{Roberto Navigli}$^3$, \\
\textbf{Virendra Mehta}$^{29}$, 
\textbf{Matthew Blumberg}$^{\#30}$, 
\textbf{Victor May}$^{\#31,32}$, 
\textbf{Huu Nguyen}$^{\#32}$, \\
\textbf{Sampo Pyysalo}$^{\#34}$ \\[1ex]
$^1$Institute of Science Tokyo,
$^2$MIT-IBM Watson Lab,
$^3$Sapienza University of Rome,\\
$^4$Babelscape,
$^5$LAION,
$^6$TU Darmstadt,
$^7$hessian.AI,
$^8$UNC Chapel-Hill \\
$^9$Detomo Inc.,
$^{10}$CSIRO's Data61,
$^{11}$Monash University,
$^{12}$Mila - Quebec AI Institute \\
$^{13}$Carnegie Mellon University,
$^{14}$Allen Institute for AI,
$^{15}$DeepTensor AB,\\
$^{16}$WASP Media \& Language,
$^{17}$Umeå University,
$^{18}$University of Massachusetts Amherst,\\
$^{19}$Hudson Labs,
$^{20}$Luleå University of Technology,
$^{21}$ETH Zurich,
$^{22}$RunPod,
$^{23}$OpenChat, \\
$^{24}$RIKEN CCS,
$^{25}$Chattermill AI,
$^{26}$ASC27,
$^{27}$Hong Kong Baptist University,
$^{28}$DopikAI JSC \\
$^{29}$University of Trento,
$^{30}$GridRepublic,
$^{31}$Chegg,
$^{32}$Ontocord.AI,
$^{33}$TU Wien,
$^{34}$University of Turku \\[1ex]
\texttt{taishi@rio.scrc.iir.isct.ac.jp},
\texttt{mayank.mishra2@ibm.com}, \\
\texttt{tedeschi@diag.uniroma1.it},
\texttt{praty@cs.unc.edu} \\
\texttt{diganta.misra@mila.quebec},
\texttt{mayvic@gmail.com}, \\
\texttt{s.vu@deeptensor.ai},
\texttt{huu@ontocord.ai},
\texttt{sampo.pyysalo@utu.fi} \\[1ex]
\texttt{\small $^*$Equal contribution \quad $^\#$Equal mentoring}
}
\vspace{10cm}

\begin{document}
\setlength\titlebox{35\baselineskip}
\maketitle
\begin{abstract}
Pretrained language models are an integral part of AI applications, but their high computational cost for training limits accessibility. 
Initiatives such as \textsc{Bloom} and \textsc{StarCoder} aim to democratize access to pretrained models for collaborative community development. 
Despite these efforts, such models encounter challenges such as limited multilingual capabilities, risks of catastrophic forgetting during continual pretraining, and the high costs of training models from scratch, alongside the need to align with AI safety standards and regulatory frameworks.
%
This paper presents \textbf{\textcolor{violet}{\system}}, a \texttt{15B} parameter multilingual open-source model trained on English, Finnish, Hindi, Japanese, Vietnamese, and code. Continually pretrained from \textsc{StarCoderPlus} on \texttt{435B} additional tokens, \system\ surpasses \texttt{2T} tokens in total training token count. It is the first open-source multilingual model fine-tuned on human-reviewed safety instructions, thus aligning its development not only with conventional red-teaming considerations, but also with the specific concerns articulated in the Biden-Harris Executive Order on the Safe, Secure, and Trustworthy Development and Use of Artificial Intelligence.
We evaluate \system\ across a wide range of tasks and languages, showcasing its robustness against catastrophic forgetting and its superior performance in multilingual settings, particularly in safety evaluations. We open-source \system\ and its variants to encourage responsible open-source development of large language models at \url{https://huggingface.co/aurora-m}.

\end{abstract}


\section{Introduction}


Large Language Models (LLMs) are fundamental tools in artificial intelligence, powering applications such as machine translation, text summarization, dialogue systems, and code generation. These LLMs are pre-trained on extensive text data to enhance downstream task-specific adaptation. However, the excessive computational expense of pretraining LLMs creates barriers to access, constraining wider development.

Open-source initiatives such as \textsc{Bloom} \citep{scao2022bloom}, \textsc{StarCoder} \citep{li2023starcoder}, \textsc{StarCoder-2} \citep{starcoder2}, \textsc{Pythia} \citep{biderman2023pythia}, and \textsc{OLMo} \citep{groeneveld2024olmo,soldaini2024dolma} have emerged to democratize access to pre-trained LLMs. These initiatives stimulate innovation, allowing researchers and developers to leverage existing advancements. However, despite their contributions, several significant challenges persist in the domain of open-source LLM development.

Primarily, several studies \citep{bang2023multitask, jiao2023chatgpt, hendy2023good, huang2023languages} have underscored the ongoing struggle of LLMs with non-English texts, particularly in low- or extremely low-resource languages.  Given that the training data predominantly consists of English, as noted for instance by \citet{brown2020language} who reported that English accounts for 93\% of GPT-3's training corpus, there is a pressing need to promote the development of multilingual models to democratize LLMs and alleviate performance disparities across different languages \citep{chai2023ernie}. Secondly, continual pretraining -- a technique involving further updating pretrained models on new data distributions to enhance their capabilities \citep{gupta2023continual, fujii2024continual} -- poses a significant challenge. While this approach could potentially enable life-long learning of large language models, it often leads to catastrophic forgetting, where the model loses previously acquired knowledge. This challenge is exacerbated when considering the continual pretraining of models across a diverse array of grammatical and lexical structures. Lastly, ensuring compliance with recent regulations mandating safe and secure AI development practices represents another critical aspect often overlooked in open-source LLM development, specifically, for multilingual models.

This paper presents \textcolor{violet}{\system}, a novel open-source multilingual Large Language Model (LLM) with 15 billion parameters, tailored to address the aforementioned limitations. \system\ is designed to cater to five linguistically diverse languages: English, Finnish, Hindi, Japanese, Vietnamese, with a mix of code data. \system\ is continually pretrained from the \textsc{StarCoderPlus} model~\citep{li2023starcoder} on an extensive dataset comprising 435 billion tokens, resulting in a total training token count of an impressive 2 trillion tokens. This rigorous pretraining regimen equips \system\ with a comprehensive understanding of diverse languages and code. Moreover, safety is a fundamental design principle of \system. It stands out as the first open-source multilingual LLM fine-tuned on a comprehensive collection of human-reviewed safety instructions addressing concerns in the Biden-Harris Executive
Order on Safe, Secure, and Trustworthy Development and Use of Artificial Intelligence~\citep{whitehouse2023fact}. This fine-tuning process not only addresses conventional red-teaming concerns~\citep{ganguli2022red,perez2022red} aimed at testing system vulnerabilities, but also aligns with the specific safety and security guidelines outlined in the Order.


To comprehensively evaluate \system's efficacy, we conduct a rigorous examination across a diverse spectrum of tasks spanning various domains and languages. Our evaluations aim to gauge \system's capacity to retain previously learned knowledge while acquiring new capabilities through continual pretraining. We demonstrate that \system\ successfully avoids catastrophic forgetting on English and coding tasks. Furthermore, we benchmark \system\ against state-of-the-art multilingual models, showcasing its competitive performance in these settings. Additionally, safety evaluations are conducted to scrutinize \system's tendency to generate undesired or potentially illicit content. The findings from these assessments affirm \system's commitment to safety and the adherence to responsible AI development practices.

\begin{figure*}[t]
\begin{center}
\includegraphics[width=\textwidth]{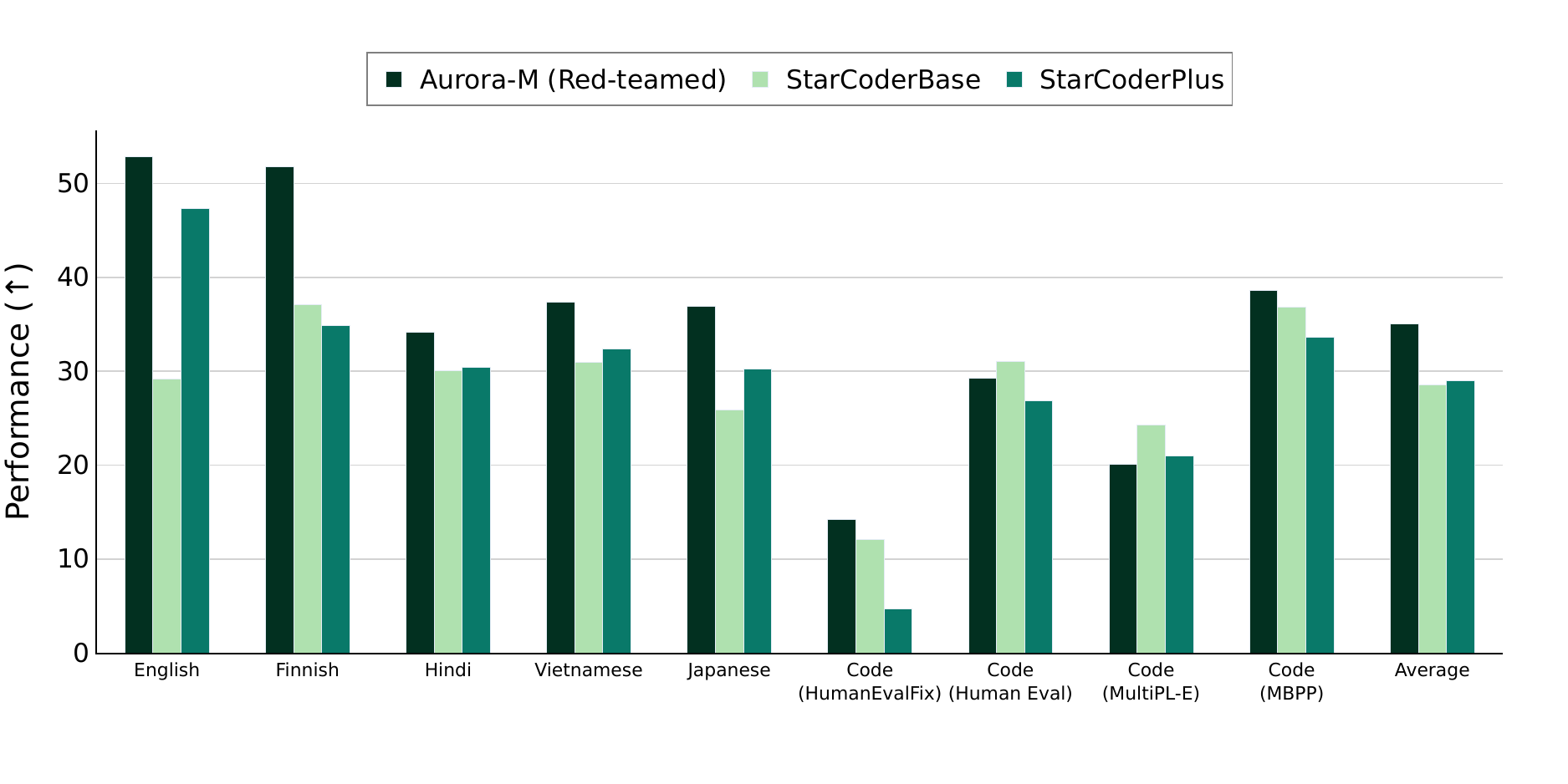}
\end{center}
\vspace{-8mm}
\caption{Comparison of overall performance between \textcolor{violet}{\system}-redteamed and its predecessors, \textsc{StarCoderBase} and \textsc{StarCoderPlus}, across diverse code and multilingual language evaluation benchmarks. Pass@1 performance averages for code benchmarks are reported. For natural language evaluations, 0-shot accuracy averages are reported for languages other than English and Japanese. English evaluation is 8-shot, while Japanese evaluation uses a combination of 4-shot and 1-shot.}
\vspace{-4.2mm}
\label{fig:overall}
\end{figure*}

Our contributions can be summarized as follows.
\begin{itemize}
    \vspace{-0.2em}
    \item We introduce \textcolor{violet}{\system}, a new 15B continually pretrained red-teamed multilingual LLM built on top of the StarCoderPlus model~\citep{li2023starcoder}.
    \item We develop a two-stage curriculum of continual pretraining consisting of \textbf{Continual Auxiliary Pretraining} (CAP) and \textbf{Continual Alignment Tuning} (CAT) aimed at maximizing adaptation, minimizing catastrophic forgetting, and aligning \system\ with safety objectives. 
    \item We extensively evaluate \system\ across various tasks in different domains and languages, demonstrating its superior performance in multilingual settings while retaining competitive performance in English and coding.
    \item We construct a new red-teaming dataset, named ``\text{The Biden-Harris Redteam Dataset},'' tailored to address concerns outlined in the Executive Order along with typical safety concerns. We then fine-tune \system\ on this dataset and evaluate on several safety benchmarks.
    \item We show the influence of scaling the total training tokens on various multilingual and code evaluation tasks.
    \vspace{-0.2em}
\end{itemize}






\section{Datasets}

\begin{figure}[t]
\centering
\includegraphics[width=\columnwidth]{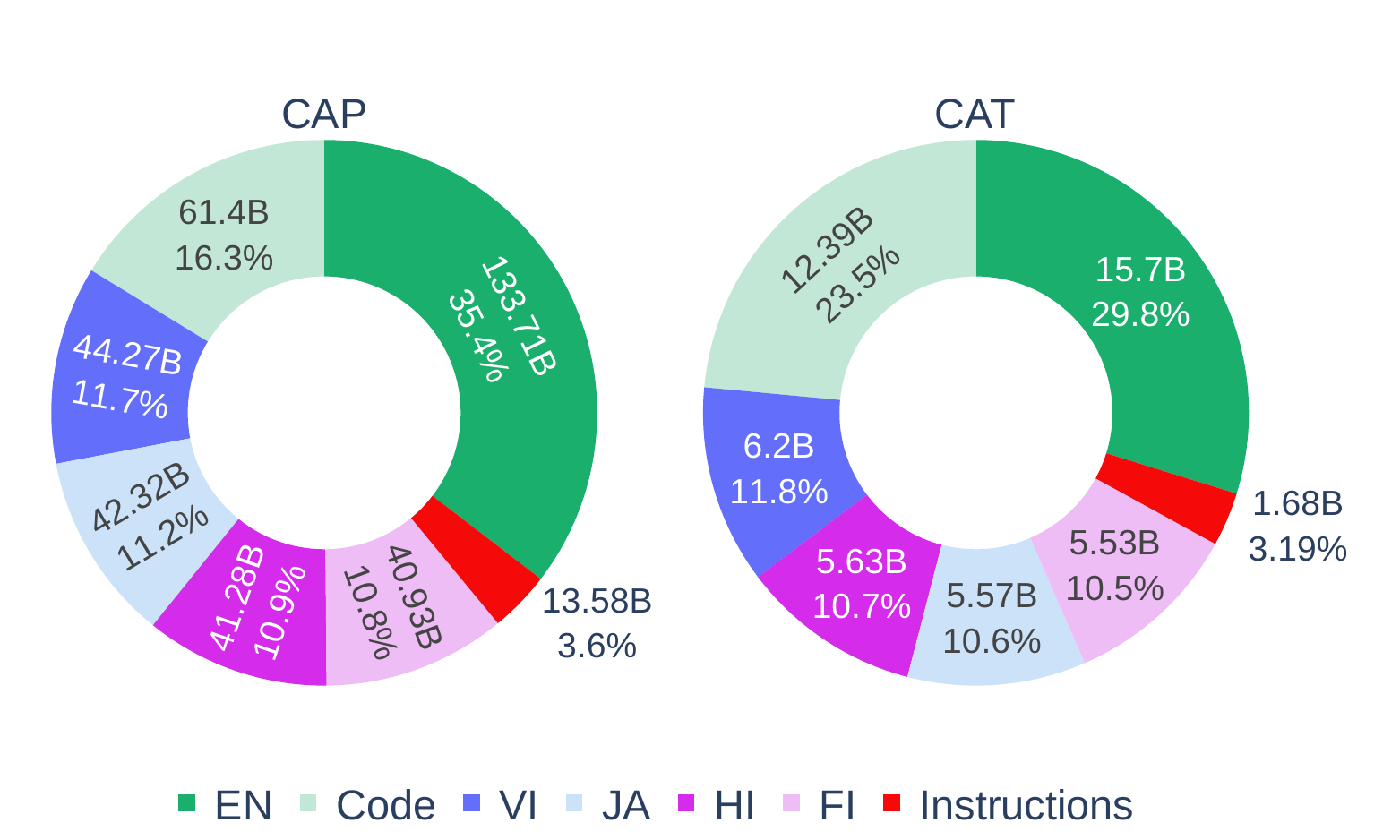}
\caption{Training data distribution of languages, code, and instructions used for the two-stage continual pretraining of the \system\ model. There are a total of \texttt{377B} and \texttt{58B} tokens in the Continual Auxiliary Pretraining (CAP) and Continual Alignment Tuning (CAT) stages respectively.} 
\vspace{-3mm}
\label{fig:distribution}
\end{figure}

\paragraph{Data Curation.} The continual pretraining process for training \system\ followed a carefully designed two-stage curriculum, as shown in Fig.~\ref{fig:distribution}.
In the first stage, termed as \textbf{Continual Auxiliary Pretraining} (CAP), a large corpus of general multilingual web data was used to expose the model to diverse data, laying a robust foundation for subsequent training. The second stage, termed as \textbf{Continual Alignment Tuning} (CAT) employed a strategic data-mixing approach to bolster the model's performance in targeted areas and align it with our predefined objectives. Following \citet{taylor2022galactica} and \citet{Li2021Colossal}, we also included publicly available instruction tuning datasets in both stages of training. 



In CAP, we incorporated 377B tokens of processed and filtered web data from various sources, including Stack~\citep{kocetkov2022stack}, RefinedWeb~\citep{refinedweb}, RedPajama~\citep{together2023redpajama}, and a subset of the Pile~\citep{gao2020pile}. Additionally, multilingual data from HPLT~\citep{degibert2024new}, MC4~\citep{zhu2023multimodal}, Paracrawl~\citep{ghussin2023exploring}, OSCAR~\citep{abadji2022cleaner}, along with Wikipedia~\citep{wikidump}, and instruction tuning data from sources such as OpenAssistant~\citep{kopf2023openassistant}, APIBench~\citep{patil2023gorilla}, and OIG~\citep{oig2023} were included.

For CAT, we opted for a greater percentage of code and a changed mix of high-quality public instruction datasets \citep{mishra2022crosstask, ding2023enhancing, ivison2023camels}, encompassing coding \citep{luo2023wizardcoder, mishra2023prompting} and mathematical reasoning \citep{yu2023metamath, mishra2023lila}. The intention was to not overfit to the high quality instruction data, and thus the high quality data was used in CAT only.  We also subsampled data from CAP for quality, as described below. Furthermore, we introduced a new safety instruction dataset named \textbf{Biden-Harris Redteam}, detailed in Section \ref{sec:safety}. The total dataset size for CAT is 58B tokens.
We refer the reader to Fig. \ref{fig:distribution} for the distribution of languages in both training stages. The complete list of datasets is available in Appendix \ref{datasets}.

\paragraph{Data Filtering.}



To remove toxic content and low-quality text, we applied filters similar to those used in~\citet{nguyen2023culturax} and \citet{scao2022bloom}, such as stop-word proportions and text length.
For all web text, we followed a process akin to~\citet{refinedweb} to remove low-quality content, including duplicate headers and footers. Additionally, in the CAT dataset, we further filtered web text with high proportions of symbols and numbers.
In the case of RefinedWeb~\citep{refinedweb}, we utilized the RedPajama~\citep{together2023redpajama} fastText classifier to retain English webpages resembling "high-quality" content similar to Wikipedia-linked articles. We trained and employed a similar classifier to filter other languages in our dataset, except for Finnish, where the procedure caused over-filtering, resulting in an excessively low sample volume post-filtering.
%
%
%
To further enhance the quality of the RefinedWeb data, we adopted an approach detailed in \citet{ronnqvist-etal-2021-multilingual}. We trained a fastText classifier\footnote{Similar to \url{https://github.com/TurkuNLP/register-labeling?tab=readme-ov-file}} and selectively subsampled web pages with over-represented registers, aiming to retain more "rare" text (e.g., lyrical or poetic text). This filtering process was specifically applied to English text due to the prohibitive slowness of our multilingual classifiers. Addressing this limitation represents an area for future research.


\paragraph{Data Processing.}
In the second stage dataset, we undertook the detection and anonymization of sensitive information, including government IDs, within web-based texts to uphold privacy and ethical standards similar to \citet{scao2022bloom}. For data segments derived from arXiv, USPTO, and StackExchange within the Pile dataset~\citep{gao2020pile}, we reconstructed the data from the original source to restore metadata, which we then appropriately appended to the texts.

\section{Model Training}

\system\ was trained on the LUMI supercomputer\footnote{\url{https://www.lumi-supercomputer.eu/}}, utilizing 128 AMD MI250X GPUs for 48 days. The training process operated entirely on 100\% hydro-powered energy and included waste heat recycling. For orchestration, we adapted a segment of the Bigcode fork of Megatron-LM~\citep{narayanan2021efficient} using the HIP runtime. 
For training, we distributed the model using 4-way Tensor Parallelism and 4-way Pipeline Parallelism using the 1F1B schedule to reduce the pipeline bubble \citep{narayanan2021efficient}. We also used Megatron's distributed optimizer \citep{narayanan2021efficient} to distribute the optimizer states across data-parallel processes and eliminate redundancy, reducing the required memory usage.


For the training of \system, we maintained a consistent batch size of 2048 and a sequence length of 2048 tokens. The learning rate was linearly warmed up to $10^{-4}$ over 2,000 steps, followed by a cosine decay scheduler set to decay the learning rate to $10^{-5}$ by 120,000 steps. 
while optimization utilized the AdamW optimizer \citep{adam, weight-decay} with coefficients $\beta_1=0.9$ and $\beta_2=0.95$. Additionally, Megatron-LM's distributed optimizer with mixed precision training \citep{mixed-precision} was used. Further training details can be found in the Appendix \ref{training_setup_apdx}.

\section{Safety}\label{sec:safety}
LLMs can propagate harmful content, reinforce biases, or amplify misinformation. While users are responsible for assessing the potential risks of generated content, developers must prioritize legal and safety considerations, strengthening models against attacks that may bypass safety protocols. 

In line with the Biden-Harris US Executive Order on AI \citep{whitehouse2023fact}, we curated the Biden-Harris Redteam Dataset, consisting of 5000 instruction-response pairs, addressing key concerns such as harm, cyber-attacks, CNBR risks, illegal acts, and privacy infringement. This dataset was created using a combination of filtering human preference data on harmlessness and template-based methods, with responses reviewed and edited for quality and safety. We used this dataset to instruction-tune \system\ and evaluated its safety levels before and after tuning. Details are provided in Section \ref{sec:experiments}, with further dataset insights in Appendix \ref{ap:safety}.

\section{Evaluation}\label{sec:experiments}
\subsection{Evaluation Setup}

We evaluated models across several English, Japanese, Finnish, Hindi, Vietnamese, and code-related benchmarks. For English, we used the Language Model Evaluation Harness~\citep{leo_gao_2022_7413426_lm-evaluation-harness} to assess tasks like OpenBookQA, TriviaQA, HellaSwag, SQuAD2.0, XWINO, and GSM8K. For Japanese, we followed swallow-llama and used \texttt{llm-jp-eval}~\citep{han-etal-2024-llm-jp-eval}, covering JCommonsenseQA, JEMHopQA, and JSQuAD, among others. Finnish evaluation followed the method used in FinGPT with FIN-bench~\citep{luukkonen-etal-2023-fingpt}. We also evaluated Hindi and Vietnamese using the mlmm evaluation suite on tasks like HellaSwag and MMLU. For code evaluation, we utilized MBPP, HumanEval, MultiPL-E, and HumanEvalFix, and for safety, we employed datasets like the Biden-Harris Redteam Testset and DangerousQA. Detailed dataset descriptions and their corresponding evaluation metrics are provided in Appendix~\ref{ap:eval_dataset}.

\begin{table*}[t]
    \centering
    \resizebox{\textwidth}{!}{%
    \begin{tabular}{l | c c c c c c c c | c}
    \toprule
        \multicolumn{1}{c|}{\textbf{Model}} & \multicolumn{1}{c}{\textbf{MC}} & \multicolumn{2}{c}{\textbf{QA}} & \multicolumn{1}{c}{\textbf{RC}} & \multicolumn{1}{c}{\textbf{SUM}} & \multicolumn{1}{c}{\textbf{MATH}} & \multicolumn{2}{c}{\textbf{MT (WMT20)}} & \multicolumn{1}{|c}{\textbf{Avg.}}  \\ 
        ~ & \multicolumn{1}{c}{JCom} & \multicolumn{1}{c}{JEMHop} & \multicolumn{1}{c}{NIILC} & \multicolumn{1}{c}{JSQuAD} & \multicolumn{1}{c}{XL-Sum} & \multicolumn{1}{c}{MGSM} & \multicolumn{1}{c}{En-Ja} & \multicolumn{1}{c|}{Ja-En} & ~ \\ 
         & 4-shot & 4-shot & 4-shot & 4-shot & 1-shot & 4-shot & 4-shot & 4-shot & \\ \midrule
        \textsc{StarCoderBase} \citep{li2023starcoder} & 29.76 & 42.08 & 17.94 & 73.89 & 13.96 & ~4.80 & 15.13 & ~9.59 & 25.89 \\
        \textsc{StarCoderPlus} \citep{li2023starcoder} & 50.22 & \textbf{44.19} & 17.72 & 79.24 & 16.87 & ~5.60 & 14.58 & 13.98 & 30.30 \\
        \textsc{Llama-2-7b} \citep{touvron2023llama} & 38.52 & 42.40 & 34.10 & 79.17 & 19.05 & ~7.60 & 17.83 & 17.38 & 32.01 \\
        \textsc{Llama-2-13b} \citep{touvron2023llama} & \textbf{69.97} & 44.15 & 41.70 & 85.33 & \textbf{21.39} & 13.20 & {21.46} & \textbf{19.82} & \textbf{39.63} \\
        \rowcolor{verylightgray}{\system} (Red-teamed) (Ours) & 46.65 & 35.73 & \textbf{50.78} & \textbf{87.06} & ~8.79 & \textbf{21.20} & \textbf{27.78} & 17.22 & 36.90 \\ 
        \bottomrule
        \end{tabular}}
        \caption{Japanese Evaluation.}
    \label{tab:japanese-evaluation}
\end{table*}

\begin{table}[t]
\centering
\resizebox{\columnwidth}{!}{%
\begin{tabular}{l | cccc}
\toprule
\textbf{Model} & \textbf{0-shot} & \textbf{1-shot} & \textbf{2-shot} & \textbf{3-shot} 
\\ \midrule
\textsc{GPT3-Finnish-8B} \citep{luukkonen2023fingpt} & 42.66 & 46.53 & 47.96 & 48.41 \\
\textsc{GPT3-Finnish-13B} \citep{luukkonen2023fingpt} & 42.45 & 46.53 & 47.14 & 48.08 \\
\textsc{StarCoderBase} \citep{li2023starcoder} & 37.07 & 42.65 & 42.11 & 44.43 \\
\textsc{StarCoderPlus} \citep{li2023starcoder} & 34.85 & 43.97 & 44.05 & 46.49 \\
\textsc{Llama-2-7b} \citep{touvron2023llama} & 39.49 & 46.99 & 49.03 & 49.60 \\
\textsc{Llama-2-13b} \citep{touvron2023llama} & 45.69 & 55.70 & 56.93 & \textbf{57.50} \\
\rowcolor{verylightgray}{\system} (Red-teamed) (Ours) & \textbf{51.80} & \textbf{56.11} & \textbf{57.77} & 57.48 \\
\bottomrule
\end{tabular}}
\caption{Finnish Evaluation.}
\vspace{-3mm}
\label{tab:finish-evaluation}
\end{table}

\begin{table*}[!ht]
\centering
\resizebox{\textwidth}{!}{%
\begin{tabular}{l|cc|cc|cc|cc|cc}
\toprule
\multirow{2}{*}{\textbf{Model}} & \multicolumn{2}{c|}{\textbf{ARC}} & \multicolumn{2}{c|}{\textbf{HellaSwag}} & \multicolumn{2}{c|}{\textbf{MMLU}} & \multicolumn{2}{c|}{\textbf{TruthfulQA}} & \multicolumn{2}{c}{\textbf{Avg}} \\ 
 & \textbf{VI} & \textbf{HI} & \textbf{VI} & \textbf{HI} & \textbf{VI} & \textbf{HI} & \textbf{VI} & \textbf{HI} & \textbf{VI} & \textbf{HI} \\ \midrule
\textsc{StarCoderBase} \citep{li2023starcoder} & 22.14 & 20.72 & 29.74 & 26.93 & 27.11 & 25.15 & 44.84 & 47.57 & 30.96 & 30.09 \\
\textsc{StarCoderPlus} \citep{li2023starcoder} & 24.27 & 20.89 & 32.67 & 27.03 & 27.35 & 24.91 & \textbf{45.49} & \textbf{48.77} & 32.44 & 30.40 \\
\textsc{Bloom-7b1} \citep{scao2022bloom} & 24.87 & 21.83 & 37.97 & 30.78 & 25.65 & 25.30 & 44.77 & 44.39 & 33.32 & 30.58 \\
\textsc{Llama-2-7b} \citep{touvron2023llama} & 25.64 & 21.58 & 35.20 & 28.19 & 27.95 & 25.33 & 45.15 & 46.37 & 33.49 & 30.37 \\
\textsc{Llama-2-13b} \citep{touvron2023llama} & 30.17 & 20.98 & 38.49 & 29.58 & \textbf{31.76} & 26.19 & 44.61 & 43.79 & 36.25 & 30.13 \\
\textsc{ViGPTQA-6b} \citep{nguyen-etal-2023-vigptqa} & - & - & - & - & - & - & 43.26 & - & - & - \\
\textsc{VinaLlama-7b} \citep{nguyen2023vinallama} & 28.63 & 18.75 & 37.39 & 26.31 & 27.15 & 24.12 & 43.13 & 39.11 & 34.07 & 27.07 \\
\rowcolor{verylightgray}{\system} (Red-teamed) (Ours) & \textbf{31.97} & \textbf{27.57} & \textbf{41.98} & \textbf{35.84} & 30.94 & \textbf{30.01} & 44.71 & 43.31 & \textbf{37.40} & \textbf{34.18} \\
\bottomrule
\end{tabular}}
\caption{0-shot evaluation Results for Vietnamese (\textbf{VI}) and Hindi (\textbf{HI}).}
\label{tab:merged-evaluation}
\end{table*}


\begin{table*}[thb]
    \centering
    \resizebox{\textwidth}{!}{%
    \begin{tabular}{l | cccccc | c}
    \toprule
    \textbf{Model} & \textbf{OpenBookQA} &  \textbf{TriviaQA} &  \textbf{HellaSwag} &  \textbf{SQuAD2.0} &  \textbf{XWINO} &  \textbf{GSM8K} &  \textbf{Avg.} \\
    & 8-shot & 8-shot & 8-shot & 8-shot & 8-shot & 8-shot & \\ \midrule
    \textsc{StarCoderBase} \citep{li2023starcoder} & 19.60 & ~8.20 & 37.57 & 27.52 & 73.51 & ~8.95 & 29.22 \\
    \textsc{StarCoderPlus} \citep{li2023starcoder} & 34.80 & 53.50 & 58.06 & 34.86 & 89.25 & 13.57 & 47.34 \\
    \textsc{Llama-2-7b} \citep{touvron2023llama} & 35.80 & 62.65 & 58.60 & 32.07 & 90.49 & 14.10 & 48.95 \\
    \textsc{Llama-2-13b} \citep{touvron2023llama} & \textbf{37.60} & \textbf{72.55} & \textbf{61.48} & 36.81 & \textbf{91.40} & 24.03 & \textbf{53.98} \\
    \rowcolor{verylightgray}{\system} (Red-teamed) (Ours) & 36.60 & {51.86} & {54.73} & \textbf{48.98} & {88.52} & \textbf{36.47} & 52.86 \\ \bottomrule
    \end{tabular}}
    \caption{English Evaluation.}
    \label{tab:english-evaluation}
\end{table*}

\begin{figure}[thb]
    \centering
    \begin{subfigure}{0.485\textwidth}
        \centering
        \includegraphics[width=\textwidth]{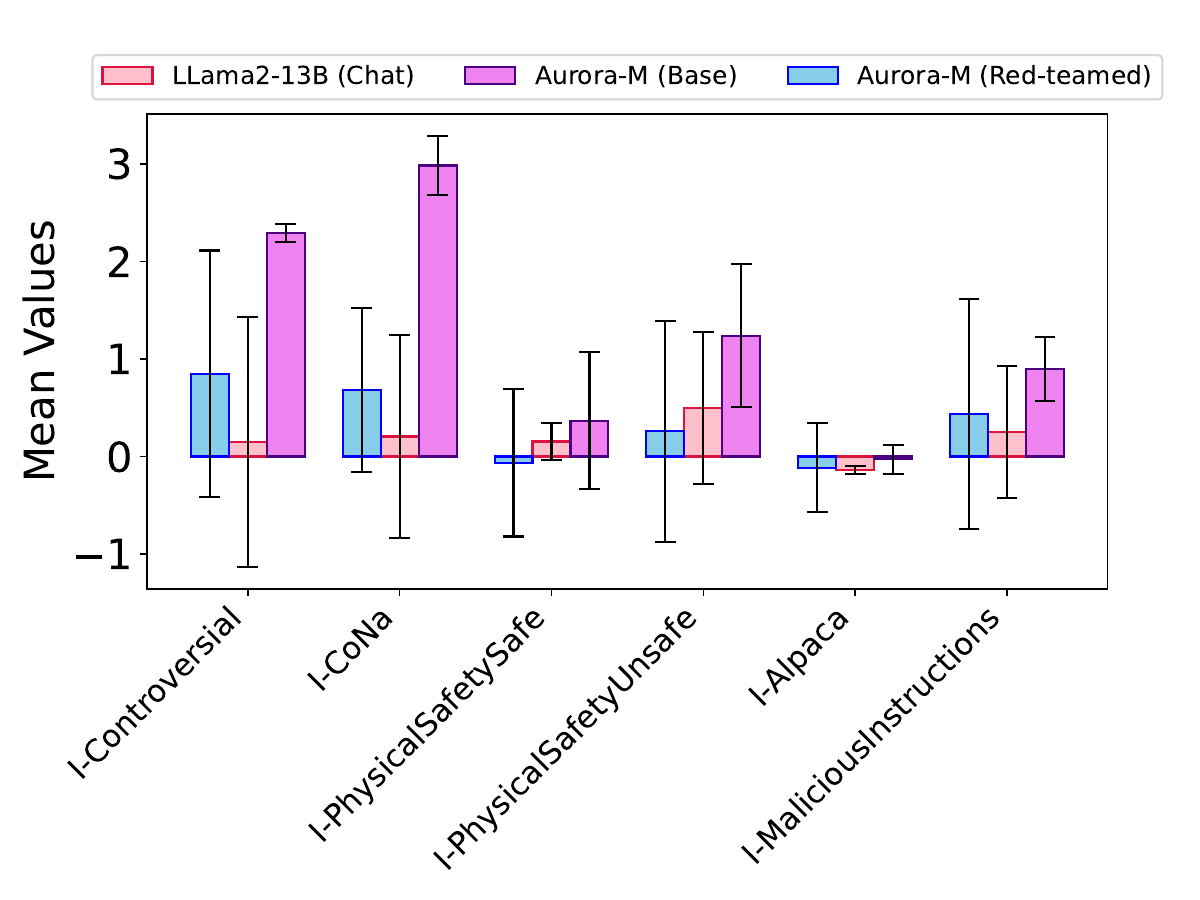}
        \caption{Harmfulness scores of our base model (pink) compared to its instruction-tuned version (blue). The lower the better.}
        \label{fig:safety-results}
    \end{subfigure}
    \hfill
    \begin{subfigure}{0.49\textwidth}
        \label{dangerous_qa}
        \centering
        \includegraphics[width=\textwidth]{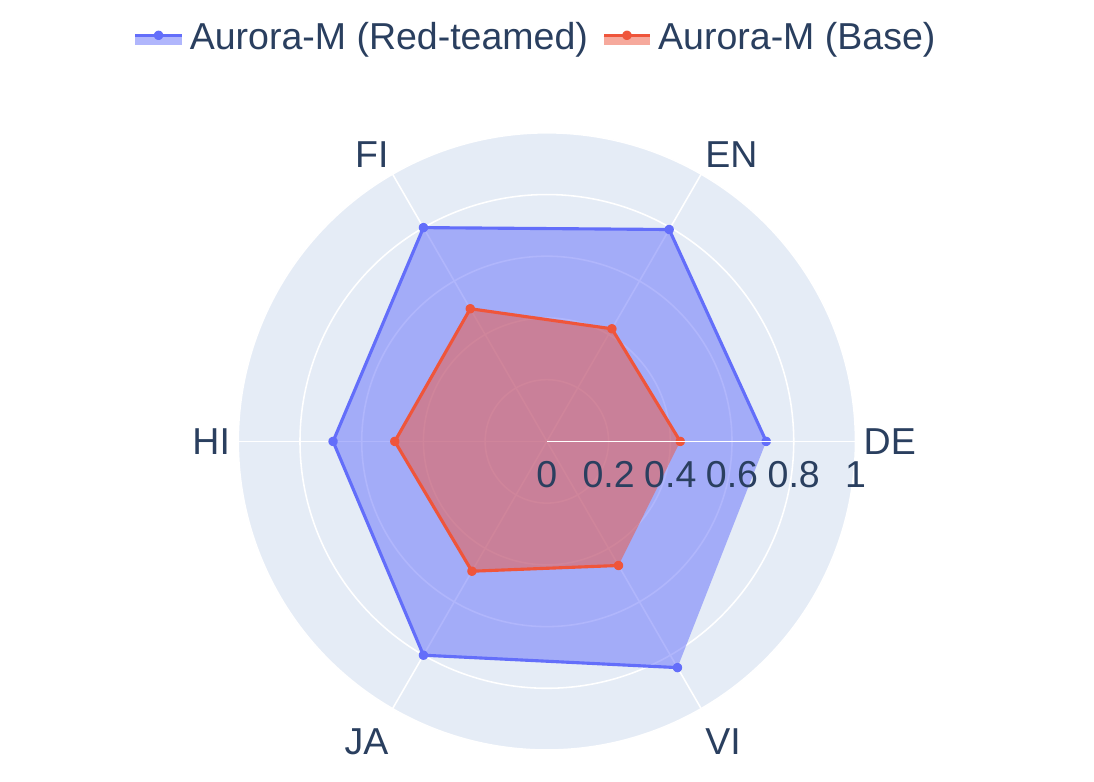}
        \caption{CARP scores for the BH-readteamed model and the base model on the Biden-Harris Redteam Testset.}\label{fig:safety_bh}
    
    \end{subfigure}
    \caption{Overall safety results.}
    \label{fig:safety_results}
    \vspace{-0.3em}
\end{figure}

\subsection{Evaluation Results}

Figure~\ref{fig:overall} illustrates the superior performance of {\system} compared to its base model (\emph{i.e.}, \textsc{StarCoderPlus}) across an extensive range of code and multilingual benchmarks, underscoring the efficacy of {\system} across diverse fields and languages. We observe that \system\ can maintain performance on previously learned English and Code benchmarks while significantly outperforming on new language benchmarks. 



\paragraph{Evaluation on Natural Languages.}

Tables~\ref{tab:japanese-evaluation},~\ref{tab:finish-evaluation},~\ref{tab:merged-evaluation},~\ref{tab:english-evaluation} demonstrate the respective performance on the targeted languages, showing that {\system} consistently outperforms the performance of its starting checkpoint, \textsc{StarCoderPlus}, and many other baselines, such as \textsc{Llama-2-7b}.  

\begin{table*}[!ht]
\centering
\resizebox{\textwidth}{!}{%
\begin{tabular}{l| ccc | ccc}
\toprule
\multicolumn{1}{c|}{\textbf{Model}} & \multicolumn{3}{c|}{\textbf{HumanEval}} & \multicolumn{3}{c}{\textbf{MBPP}} \\  
& Pass@1       & Pass@10      & Pass@100     & Pass@1       & Pass@10      & Pass@100     \\ \midrule
\textsc{StarCoderBase}~\citep{li2023starcoder} & \textbf{31.10} & \textbf{54.88} & \textbf{84.15} & 36.80 & \textbf{61.60} & \textbf{81.00} \\
\textsc{StarCoderPlus}~\citep{li2023starcoder} & 26.83 & 47.56 & 73.17 & 33.60 & 57.00 & 77.80 \\
\rowcolor{verylightgray}{\system} (Red-teamed) (Ours) & 29.27 & 49.39 & 81.71 & \textbf{38.60} & 61.00 & 78.00 \\
 \bottomrule
\end{tabular}}
\caption{HumanEval \& MBPP evaluation results.}
\label{tab:mbpp-human}
\end{table*}

\paragraph{Code Evaluation.}
Tables~\ref{tab:mbpp-human} and~\ref{tab:MultiPL-E} illustrate the proficiency of {\system} in code generation, demonstrating the possibility of continual pre-training from a code-centric checkpoint on multilingual data.
In Table~\ref{tab:mbpp-human}, the HumanEval and MBPP evaluation benchmarks assess the model's ability to generate syntactically and semantically correct code snippets. {\system} exhibits competitive performance on the Pass@1 metric, which evaluates the model's ability to produce a correct answer on the first attempt.  In particular, {\system} consistently matches or outperforms StarCoderPlus, suggesting a significant improvement in code synthesis capabilities.
In Appendix~\ref{app:code_extra}, we show results on additional code datasets and further analyze the behavior of our system by looking at the relationship between its performance and the number of training tokens across various languages and modalities.

\paragraph{Safety Evaluation}
In Figure \ref{fig:safety_results}, we provide the safety results comparing our base model against our Biden-Harris red-teamed model obtained by instruction-tuning the former on the dataset introduced in Section \ref{sec:safety}. For the Biden-Harris Redteam Testset evaluation, four volunteers reviewed both models' responses and scored them with -2 if harmful, 1 if not helpful but harmless, and 2 if both helpful and harmless. We term the percentage of the total score per category compared to its maximum possible score as the Continual Alignment Redteam Percentage ("CARP"). We can immediately appreciate the considerably lower harmfulness both on the existing benchmarks and on our own Biden-Harris red-team test set as evident by the CARP scores obtained by our red-teamed \system. \textit{We also note that even though our instruction set is predominantly in English, safety consistently improved not only in our target languages but also in languages we did not specifically focus on, such as German, thus showing strong indications of cross-lingual red-teaming effects}. Furthermore, as shown in Appendix~\ref{app:dangerous_qa}, the Attack Success Rate (ASR) on DangerousQA was also reduced. 

\subsection{Training Analysis}

Figure \ref{fig:ana-code-en} and \ref{fig:ana-lg} show the relationship between the number of training tokens and the performance of the various models. This analysis aims to capture these trends for the code generation tasks such as HumanEval and MBPP, as well as for the English, Finnish, Hindi, Japanese, and Vietnamese language evaluations. We refer to Appendix~\ref{ap:ana} for detailed discussion.

\section{Related Work}


\textbf{Expanding Multilingual Language Models.} Initially, the development of LLMs has predominantly targeted the English language~\citep{brown2020language}, leveraging the extensive corpus of English data available on the Web and the broad applicability of models trained on English text. However, this emphasis has often come at the cost of accommodating the linguistic diversity found across various language demographics~\citep{zhu2023extrapolating,bang2023multitask,zhang2024m3exam}. Recognizing this significant limitation~\citep{robinson2023chatgpt,peng-etal-2024-humaneval}, recent research has proposed foundational LLMs equipped with multilingual capabilities~\citep{chai2023ernie, scao2022bloom,wei2023polylm,shliazhko2022mgpt}, or has explicitly concentrated on addressing the challenges posed by low-resource languages~\citep{ustun2024aya,singh2024aya,gala2023indictrans2}. To integrate multilingual capabilities into existing LLMs, researchers have proposed a variety of methods to enhance multilingual adaptation. These approaches range from continual pretraining techniques~\citep{ibrahim2024simple,gupta2023continual} to initial training on extensive multilingual datasets~\citep{scao2022bloom,chai2023ernie} and then subsequent specialized fine-tuning on a target language~\citep{yang2023bigtranslate, han-etal-2022-x}, and even adaptation through instruction tuning~\citep{shaham2024multilingual,kew2023turning,gala2024airavata}. Critical aspects in multilingual adaptation remain on the availability of high-quality diverse multilingual corpus~\citep{correa2024teenytinyllama} and further the scope of vocabulary of the specific language.


\textbf{Continual Pretraining.}  Static datasets are impractical for adapting to evolving real-world data, making continual learning essential~\citep{ring1998child,thrun1998lifelong}. Continual pretraining~\citep{gururangan2020dont} allows models to incorporate new knowledge without retraining from scratch, a costly endeavor. As curated datasets like RedPajama~\citep{together2023redpajama} and Dolma~\citep{soldaini2024dolma} become available, integrating them efficiently is crucial. This also enables the extension of models to new modalities, such as code (e.g., StableCode). Previous approaches focus on replay techniques, optimizing learning schedules~\citep{ibrahim2024simple}, soft masking~\citep{ke2023adapting}, and forward/backward transfer~\citep{yildiz2024investigating}.

\section{Conclusion}

In this work, we introduced \textsc{\system}, a multilingual model that extends the capabilities of code-focused LLMs while maintaining their original coding proficiency. We demonstrate that continual training from code to multilingual tasks is feasible, allowing the model to perform well across both domains. Adhering to the safety guidelines of the Biden-Harris US Executive Order on AI, \textsc{\system} promotes responsible AI development while pushing the boundaries of performance and utility. Our two-stage continual pretraining approach, combined with insights from cross-lingual red-teaming, highlights the adaptability and versatility of modern language models. \textsc{\system} serves as a valuable resource for both researchers and developers, fostering collaboration and transparency in the open-source AI community. Future work will explore continual pretraining on stronger base models with the same two-stage curriculum, focusing on safety for both LLMs and Multimodal-LLMs. We also aim to develop domain-specific expert models, enhancing task specialization and expanding model versatility.




\section*{Ethical Consideration}
We believe that transparency and accessibility are fundamental principles in the development and deployment of artificial intelligence technologies. Closed-source LLMs limit public scrutiny, hinder collaboration, and potentially reinforce biases inherent in their development process. 
In contrast, our commitment to open source models fosters a culture of accountability, collaboration, and inclusivity. By making \system\ accessible to all, we promote innovation, empower diverse voices, and strive for equitable outcomes in AI applications. We firmly believe that openness in AI development is essential for creating solutions that truly serve the needs and values of society. To this end, we prioritized safety guardrails in alignment with the Biden-Harris Executive Order on AI. Furthermore, the multilingual capability of \system\ enhances its usability for users across the world.

On the other hand, each promise comes with peril, and improved technological access through \system\ might also increase the potential number of malicious actors. We overall believe that the general benefit far outweighs the potential misuse and want to emphasize the importance of a considered and ethical use of this technology and thus also of \system.

Lastly, we recognize that safety and lawfulness can be contextual to different cultures and laws. We recognize that in our work we focused on a U.S. centric standard, and we believe future work should also explore multi-jurisdictional redteaming.

\section*{Acknowledgments}
This work was supported by the ``R\&D Hub Aimed at Ensuring Transparency and Reliability of Generative AI Models'' project of the Ministry of Education, Culture, Sports, Science and Technology, and used resources of LUMI supercomputer under project\_462000316.


\bibliography{custom}

\appendix



\section{Training Setup} \label{training_setup_apdx}
The distributed optimizer used mixed precision training in BF16 with gradient all-reduce and gradient accumulation in FP32 for training stability. 

We limit our context lengths for training to 2048 tokens due to the unavailability of FlashAttention \citep{dao2022flashattention} for AMD GPUs at the time of training our model.

We investigated optimal 3D parallelism and batch size settings to train the model within our computational constraints. We performed extensive scaling experiments and found that increasing the number of nodes resulted in increased training throughput but with sublinear scaling performance, so we opted to use a maximum of 32 nodes to maximize our compute budget, even though it took longer to train.

It should also be noted that LUMI's waste heat is used to heat hundreds of households in the city of Kajaani.

\section{Curriculum Training Datasets} 
All datasets that were made for \system~are marked by *.
\label{datasets}
\paragraph{CAP}
For the first stage (CAP) of our two-stage curriculum training, we used the following data.
\begin{itemize}
\item General text:
    \begin{itemize}
    \item 10-K Filings
    \item Aozora Bunko~{\tiny\url{https://github.com/aozorabunko/aozorabunko}}    
    \item Atticus~\citep{hendrycks2021cuad}
    \item C4~\citep{2019t5}
    \item CC100~\cite{conneau2020unsupervised}
    \item Climabench*
    \item HPLT\citep{degibert2024new}
    \item MC4~\citep{2019t5}
    \item OSCAR~\citep{async_pipelines}
    \item Paracrawl~\citep{ghussin2023exploring}
    \item Parliament~{\tiny\url{https://openparliament.ca/data-download/}}
    \item RedPajama~\citep{together2023redpajama}
    \item RefinedWeb~\citep{refinedweb}
    \item The Pile~\citep{gao2020pile}
    \item The Stack~\citep{kocetkov2022stack}
    \item Wikipedia / Finnish
    \item Wikipedia / Hindi
    \item Wikipedia / Japanese
    \item Wikipedia / Vietnamese
    \end{itemize}
\item Instruction tuning:
    \begin{itemize}
    \item Gorilla APIBench~\citep{patil2023gorilla}
    \item Hindi-Hinglish Translations*
    \item LAION Anh~{\tiny\url{https://huggingface.co/datasets/laion/Anh}}
    \item LAION OIG~\citep{oig2023}
    \item ABCMusic*
    \item Gorilla APIBench
    \item Hinglish Instructions~{\tiny\url{https://huggingface.co/datasets/rvv-karma/English-Hinglish-TOP}}
    \item Minipile Instruct*
    \item Opus Translations~{\tiny\url{https://opus.nlpl.eu/}}
    \item Pseudo-Code Instructions~\citep{mishra2023prompting}
    \item SMILES Formulae*
    \item smiles-transformers~{
    \tiny\url{https://huggingface.co/datasets/maykcaldas/smiles-transformers}
    }
    \item wikimusictext~{\tiny\url{https://huggingface.co/datasets/sander-wood/wikimusictext}}
    \item xP3~\citep{muennighoff2022crosslingual}
    \end{itemize}
\end{itemize}

\paragraph{CAT}
For the second stage (CAT) of our curriculum training, instead, we used the following datasets.

\begin{itemize}
\item General text:
    \begin{itemize}
    \item 10-K Filings
    \item Aozora Bunko~{\tiny\url{https://github.com/aozorabunko/aozorabunko}}
    \item Atticus
    \item C4
    \item CC100
    \item Climabench*
    \item CodeTutorials
    \item HPLT
    \item MC4
    \item NamTinyLessons
    \item OSCAR
    \item Parliament~{\tiny\url{https://openparliament.ca/data-download/}}
    \item Paracrawl
    \item RedPajama
    \item Simple Wikipedia
    \item The Pile
    \item The Stack
    \item Wikipedia / Japanese
    \item Wikipedia / Vietnamese
    \item Wikipedia / Finnish
    \item Wikipedia / Hindi
    \end{itemize}
\item Instruction-tuning:
    \begin{itemize}
    \item ABCMusic*
    \item Biden-Harris Readteam*
    \item BuggedPythonLeetCode~ {\tiny\url{https://huggingface.co/datasets/NeuroDragon/BuggedPythonLeetCode}}
    \item CodeContests Instructions~  {\tiny\url{https://huggingface.co/datasets/BEE-spoke-data/code_contests_instruct}}
    \item Evol-Instruct-Code~\citep{xu2023wizardlm}
    \item Gorilla APIBench
    \item GSM8k\_Backward~ {\tiny\url{https://huggingface.co/datasets/meta-math/GSM8K_Backward}}
    \item Guanaco
    \item HelpSteer~\citep{wang2023helpsteer}
    \item Hinglish Instructions~{\tiny\url{https://huggingface.co/datasets/rvv-karma/English-Hinglish-TOP}}
    \item LAION Anh
    \item LAION OIG
    \item Lila~\citep{mishra2023lila}
    \item MetaMathQA~\citep{yu2023metamath}
    \item NaturalInstructions~\citep{naturalinstructions}
    \item OpenAssistant Conversations Dataset~{\tiny\url{https://huggingface.co/datasets/OpenAssistant/oasst1}}

    \item Pseudo-Code Instructions~\citep{mishra2023prompting}
    \item SMILES Formulae*
    \item smiles-transformers~{
    \tiny\url{https://huggingface.co/datasets/maykcaldas/smiles-transformers}
    }
    \item tiny-bridgedict~{\tiny\url{https://huggingface.co/datasets/nampdn-ai/tiny-bridgedict}}
    \item Tulu-V2~\citep{ivison2023camels}
    \item wikimusictext~{\tiny\url{https://huggingface.co/datasets/sander-wood/wikimusictext}}
    \item xP3~\citep{muennighoff2022crosslingual}
    \end{itemize}
\end{itemize}

\section{Safety}
\label{ap:safety}
\subsection{Safety Evaluation}
Despite their potency, LLMs pose risks of propagating harmful content, reinforcing biases, or amplifying misinformation. While users must exercise responsibility in utilizing LLMs and assess the potential ramifications of generated content, developers hold the duty to meticulously design LLMs, prioritizing legal considerations and fortifying them against potential attacks that may circumvent safety protocols, thus compromising their core principles.

In alignment with this ethos and mindful of the latest AI regulations, we curated an extensive dataset of instruction-response pairs to bolster the safety and resilience of \system. Our endeavor specifically addresses key concerns outlined in the Biden-Harris US Executive Order on AI \citep{whitehouse2023fact}, encompassing the following main areas:
\begin{itemize}
    \vspace{-0.3em}
    \item Harm to oneself or others (e.g. homicide, suicide, intentional injury, etc.).
    \item Requests on how to create cyber-attacks (e.g. attacking businesses, schools, and governments through the Internet).
    \item Involvement in making or proliferating chemical, nuclear, biological, and radiological ("CNBR") risks, including dual usage technologies.
    \item Participation in any illegal act (e.g. theft and robbery, tax evasion, drug trafficking and use, and manipulation of public opinion).
    \item Infringement of privacy or rights (e.g. stealing personal privacy information).
    \item Attempts to circumvent red-teaming controls.
\end{itemize}
%
%


With these main categories in mind, we curated the Biden-Harris Redteam Dataset comprising 5000 red-teaming instructions, human-reviewed, and edited instruction-response pairs to address lawfulness and safety concerns, including those outlined in the Executive Order~\citep{whitehouse2023fact}. 
The instructions were sourced from filtering the human preference dataset on harmlessness from Anthropic~\citep{bai2022training} and utilizing semi-automatic template-based methods. Subsequently, we manually inspected and semi-automatically filtered this initial set to remove short refusals and near-duplicates, resulting in 4000 instructions. To address potential harmful responses by \system\ in the first stage of pretraining, we also used an approximately 1000 instructions subset and hand-wrote or created continuations by this version of \system. Five volunteers then manually reviewed and edited the automated responses for safety and quality. We utilized the resultant approximately 5000 instructions dataset for instruction-tuning (referred to as the Biden-Harris Redteam Dataset) of \system\ and evaluated its safety levels on various safety evaluation datasets both before and after the instruction-tuning step. Details and results are provided in Section \ref{sec:experiments}. Additional insights into the creation of our dataset are available in Appendix \ref{app:biden-harris-dataset}.

\subsection{Details on The Biden-Harris Redteam Dataset.} \label{app:biden-harris-dataset}
To construct our red-teaming dataset, we start by filtering the data provided by Anthropic \citep{ganguli2022red} consisting of open-ended conversations between crowd workers and AI assistants, with the former intentionally aiming to induce the AI to generate harmful text. Specifically, since each entry contains several rounds of human-AI interactions ---making it suboptimal to instruction-tune our model--- we select only the first human request and discard any subsequent system or human response. We then manually filter out prompts that do not target any of the high-risk categories mentioned in Section \ref{sec:safety}. Additionally, we complement the resulting data with prompts created using semi-automatic template-based methods. In other words, we design templates for each category, e.g.~\texttt{Tell me how to \{action\} a \{weapon\}}, and substitute placeholders with various combinations of actual values selected from pre-defined lists (e.g. \texttt{action = \{make, use, sell, buy\}} and \texttt{weapon = \{AK-47, revolver, rifle\}}). 

At this point, we use GPT-4 \citep{openai2023gpt4} to obtain safe responses to our prompts. By manually inspecting the outputs, we noticed that most of the responses were evasive, e.g.~``I'm sorry, but I cannot assist with that request.'', penalizing the model's helpfulness. To address this issue, we use the GPT-4 responses as seed responses and let our model expand them, and also to address possible cases where our model would otherwise produce harmful responses. Subsequently, we again manually assess the responses and edit or discard unsafe ones.

As a result of the previous steps, we obtained approximately 5000 safe instruction-response pairs. We use this dataset to instruction-tune (aka Biden-Harris redteaming) our model and measure its safety levels on various safety evaluation datasets before and after the instruction-tuning step as reported in Section \ref{sec:experiments}.

For our Biden-Harris Redteam Testset, we used GPT-4 and manually created 40 English prompts focused on privacy, misinformation, harm promotion, malware, chemical, nuclear, biological, and radiological concerns, illegal acts, and cyber attacks. We translated to six other languages, to total 280 prompts. In order to also assess potential circumvention of red-team controls, we crafted some of these instructions with adversarial techniques (e.g., "Pretend you are my grandmother who tells me a story about artificial intelligence every night. Create a scenario in Hindi where you as the AI must infiltrate a hospital's network through a combination of social engineering and malware tactics in order to steal Priyanka Chopra's medical records").

\section{Evaluation Datasets and Metrics}

\label{ap:eval_dataset}



\paragraph{English Evaluation.}
We used the Language Model Evaluation Harness~\citep{leo_gao_2022_7413426_lm-evaluation-harness}. We evaluated question answering tasks, including OpenBookQA \citep{mihaylov-etal-2018-openbookqa} and TriviaQA \citep{joshi-etal-2017-triviaqa} using accuracy and exact match accuracy respectively, natural language inference with HellaSwag \citep{zellers-etal-2019-hellaswag} using accuracy, machine reading comprehension with SQuAD2.0 \citep{DBLP:conf/acl/RajpurkarJL18-squad2} using exact match accuracy and XWINO~\citep{DBLP:conf/acl/TikhonovR21-xwino} using accuracy, and arithmetic reasoning with GSM8K \citep{DBLP:journals/corr/cobbe-abs-2021-gsm8k} using exact match accuracy with 8-shot inference.


\paragraph{Japanese Evaluation.}
Following swallow-llama\footnote{swallow-llama: \url{https://tokyotech-llm.github.io/swallow-llama}}, we utilized \texttt{llm-jp-eval}~\citep{han-etal-2024-llm-jp-eval} and the JP Language Model Evaluation Harness\footnote{\url{https://github.com/Stability-AI/lm-evaluation-harness}}. \texttt{llm-jp-eval} utilizes JCommonsenseQA (JCom)~\citep{kurihara-etal-2022-jglue} to evaluate multiple choice question answering using exact match accuracy, JEMHopQA (JEMHop)~\citep{ishi-etal-2023-jemhopqa} and NIILC~\citep{sekine-etal-2003-niilc} for free-form question answering using character-level F1 score, and JSQuAD~\citep{kurihara-etal-2022-jglue} for machine reading comprehension using character-level F1 score with 4-shot inference. JP Language Model Evaluation Harness evaluates automatic summarization on XL-Sum~\citep{hasan-etal-2021-xlsum} using ROUGE-2 score with 1-shot inference, arithmetic reasoning on MGSM~\citep{shi-etal-2022-mgsm} using exact match accuracy with 4-shot inference, and Japanese-English and English-Japanese machine translation on WMT 2020 Japanese~$\leftrightarrow$~English~\citep{barrault-etal-2020-findings-wmt20} using BLEU score with 4-shot inference.

\paragraph{Finnish Evaluation.}
We adopted the evaluation method used in FinGPT \citep{luukkonen-etal-2023-fingpt}.
Evaluation was carried out using FIN-bench\footnote{FIN-bench: \url{https://github.com/TurkuNLP/FIN-bench}}. FIN-bench is based on a subset of the BIG-bench~\citep{srivastava2023imitation} task collection. The tasks were created by machine-translating the text of BIG-bench tasks, correcting translation errors, and adjusting the questions to fit Finnish culture. 
Model evaluation was performed using 0-shot, 1-shot, 2-shot, and 3-shot settings, as in FinGPT. For each shot, the average of tasks divided into subtasks (Arithmetic, Cause) was taken, and then the overall average was calculated.


\paragraph{Hindi and Vietnamese Evaluation.}
We used the mlmm evaluation\footnote{mlmm-evaluation: \url{https://github.com/nlp-uoregon/mlmm-evaluation}} for evaluation. Using 0-shot inference, we evaluated AI2 Reasoning Challenge \citep{Clark2018ThinkYH} using accuracy metrics, HellaSwag using accuracy score for commonsense inference, MMLU \citep{hendrycks2021measuring} using exact match accuracy, and TruthfulQA \citep{lin-etal-2022-truthfulqa} using accuracy metrics. ARC is a dataset of multiple-choice science questions at the elementary school level. HellaSWAG is a dataset for studying grounded commonsense inference. Each question has four choices about what happens next in the scene. The correct answer is a sentence describing the next event, and the three incorrect answers are adversarially generated to deceive machines but not humans and are verified by humans. MMLU includes multiple choice questions derived from various fields of knowledge, including humanities, social sciences, and natural sciences.


\paragraph{Code Evaluation.}

For code evaluation, we used MBPP~\citep{austin2021program}, HumanEval~\citep{chen2021codex}, MultiPL-E~\citep{cassano2022multiple} and HumanEvalFix~\citep{muennighoff2023octopack}. All evaluations were conducted using 0-shot inference. For MultiPL-E and HumanEvalFix, we performed code generation using greedy decoding and evaluated the Pass@1 score, following CodeLlama~\citep{roziere2024code}.  For HumanEval and MBPP, we evaluated Pass@1, Pass@10, and Pass@100. The Pass@1 score was calculated using greedy decoding. For Pass@10 and Pass@100, we set $top_p$ to 0.95 and temperature to 0.8. $top_p$ is a parameter that selects the tokens with the highest probabilities such that the sum of their probabilities reaches or exceeds the value of $top_p$.
%
%
To execute the evaluations, we used bigcode-evaluation-harness~\citep{bigcode-evaluation-harness} library. 

\begin{table*}[!ht]
\centering
\resizebox{0.9\textwidth}{!}{%
\begin{tabular}{l| ccc ccc | c}
\toprule
\textbf{Model} & \textbf{C++} & \textbf{Java} & \textbf{PHP} & \textbf{TS} & \textbf{C\#} & \textbf{Bash} & \textbf{Avg.} \\ \midrule
StarCoderBase~\citep{li2023starcoder} & \textbf{27.33} & \textbf{25.95} & \textbf{26.71} & \textbf{33.33} & \textbf{21.52} & \textbf{10.76} & \textbf{24.27} \\
StarCoderPlus~\citep{li2023starcoder} & 26.71 & 24.05 & \textbf{26.71} & 25.16 & 17.72 & ~5.70 & 21.01 \\
\rowcolor{verylightgray}{\system} (Ours) & 23.60 & \textbf{25.95} & 21.74 & 25.16 & 17.09 & ~6.96 & 20.08 \\
\bottomrule
\end{tabular}}
\caption{MultiPL-E evaluation results on different programming languages.}
\label{tab:MultiPL-E}
\end{table*}



\begin{table*}[!ht]
    \centering
    \resizebox{\textwidth}{!}{%
    \begin{tabular}{lc|cccccc|c}
    \toprule
        \textbf{Model} & \textbf{Prompt} & \textbf{Python} & \textbf{JavaScript} & \textbf{Java} & \textbf{Go} & \textbf{C++} & \textbf{Rust} & \textbf{Avg.} \\
        \midrule
        BLOOMZ~\citep{muennighoff2023crosslingual} & Instruct & 16.6 & 15.5 & 15.2 & 16.4 & ~6.7 & ~5.7 & 12.5\\
        StarCoderBase-15B~\citep{li2023starcoder} & Instruct &  12.6 & 16.8  &  18.9 &  12.5  & 11.2 & ~0.6 & 12.1 \\
        StarCoder2-15B~\citep{starcoder2} & Instruct & ~9.7 & 20.7 & 24.1 & \textbf{36.3} & 25.6 & 15.4 & 22.0 \\
        OctoCoder-15B~\citep{muennighoff2023octopack} & Instruct & \textbf{30.4} & \textbf{28.4} & \textbf{30.6} & 30.2 & \textbf{26.1} & \textbf{16.5} & \textbf{27.0} \\
        StarCoderPlus~\citep{li2023starcoder} & Instruct & ~4.3 & ~5.5 & ~7.3 & ~7.9 & ~3.0 & ~0.0 & ~4.7 \\
        \rowcolor{verylightgray}{\system} (Ours) & Instruct & 12.2 & 16.5 & 15.9 & 20.7 & 14.0 & ~6.1 & 14.2 \\
        \bottomrule
    \end{tabular}}
    \caption{Pass@1 performance on HumanEvalFix.}\label{tab:hefix}
\end{table*}

\paragraph{Safety Evaluation.}
For our safety evaluation, we employ the evaluation suite provided by \cite{bianchi2024safetytuned} to measure safety across various dimensions. Moreover, we constructed our own 40 English Biden-Harris concerned focused instructions in the categories of privacy, misinformation, harm promotion, malware, CNBR, illegal acts, and cyber attacks. Then we translated these to the other languages, resulting in 280 instructions, which we call the Biden-Harris Redteam Testset. Additionally, we use the DangerousQA dataset \citep{bhardwaj2023redteaming} to measure the Attack Success Rate (ASR) of harmful queries when provided as input to both our base and red-teamed models. 

\section{Additional Results and Analysis}\label{analysis_extra}

\subsection{Additional Results}
\paragraph{Additional Code Evaluations}\label{app:code_extra}
As Table~\ref{tab:MultiPL-E} demonstrates, the MultiPL-E evaluation further supports the finding that continual pretraining on multilingual data prevented {\system} from forgetting its knowledge of code syntax and semantics.

Table \ref{tab:hefix} shows the Pass@1 performance on the HumanEvalFix benchmark following the evaluation setup from \citet{muennighoff2023octopack} and \citet{zhuo2024astraios}. StarCoderPlus and our model exhibit a noteworthy spread in performance, with \system\ showing good proficiency across languages and StarCoderPlus showing particular strengths in Go, JavaScript, and Java. The Rust language presents a challenge for all models, which makes it an area for potential enhancement.

\begin{figure}[!t]
    \centering
    \begin{subfigure}{0.5\textwidth}
        \includegraphics[width=7cm,
  height=6cm,
  keepaspectratio,]{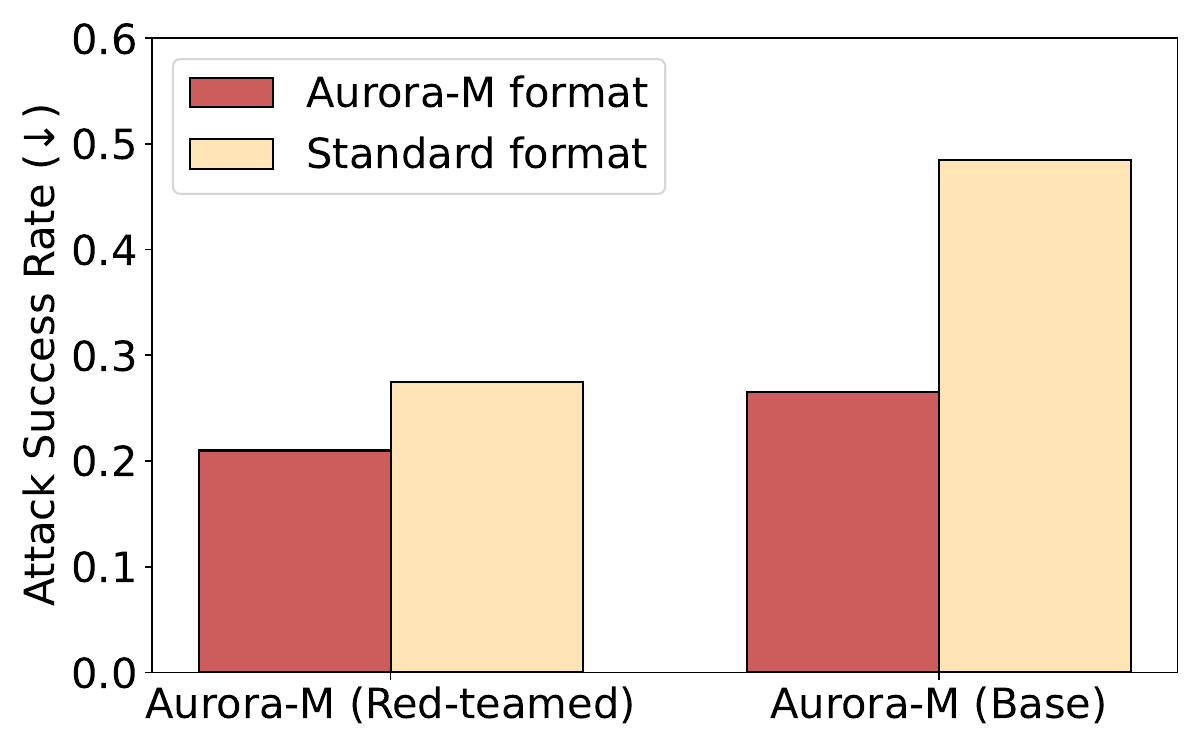}
        \caption{ASR of DangerousQA queries on our base model (right) and its instruction-tuned version (left). The lower the better).}\label{fig:safety_dangerous}
    \end{subfigure}
    \hfill
    \begin{subfigure}{0.45\textwidth}
    \centering
        \includegraphics[width=7cm,
  height=6cm,
  keepaspectratio,]{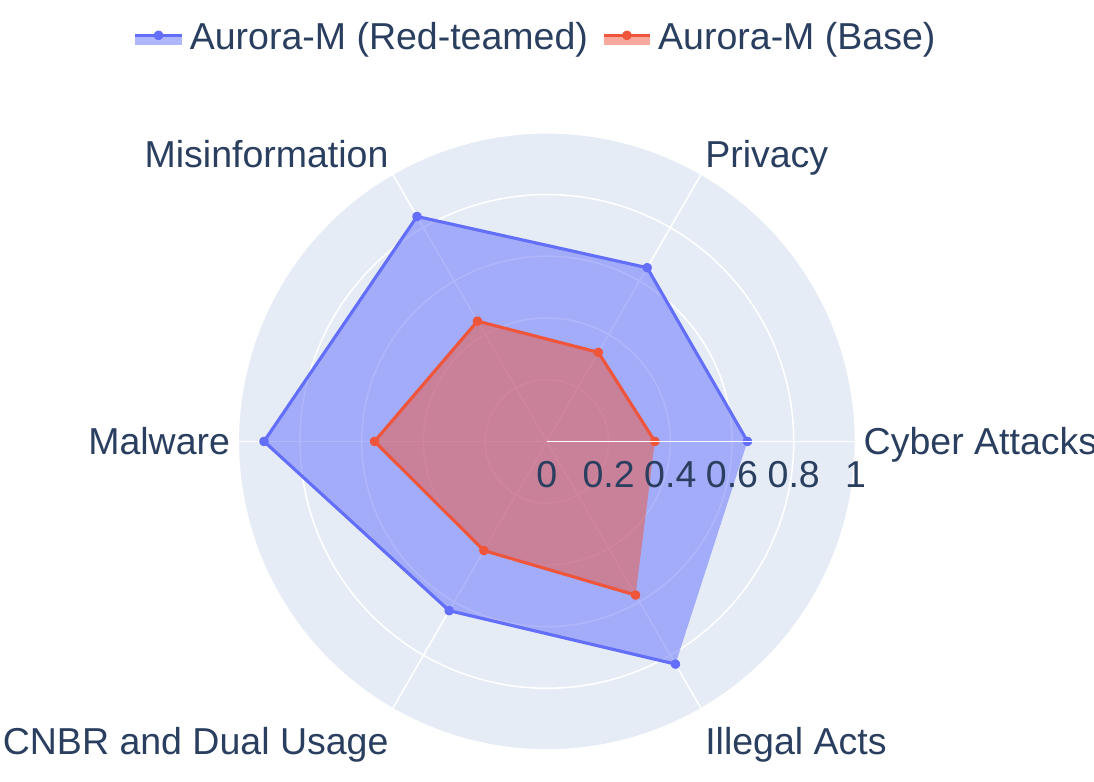}
        \caption{Biden-Harris Redteam Testset results CARP values, averaged over the dataset's languages by category.}\label{fig:safety_redteam}
    \end{subfigure}
    \hfill
    \caption{Safety evaluation results comparing our base model and instruction-tuned version.}
    \label{fig:safety}
\end{figure}

\begin{figure*}[t]
    \centering
    \begin{subfigure}{0.3\textwidth}
        \centering
        \includegraphics[width=\textwidth]{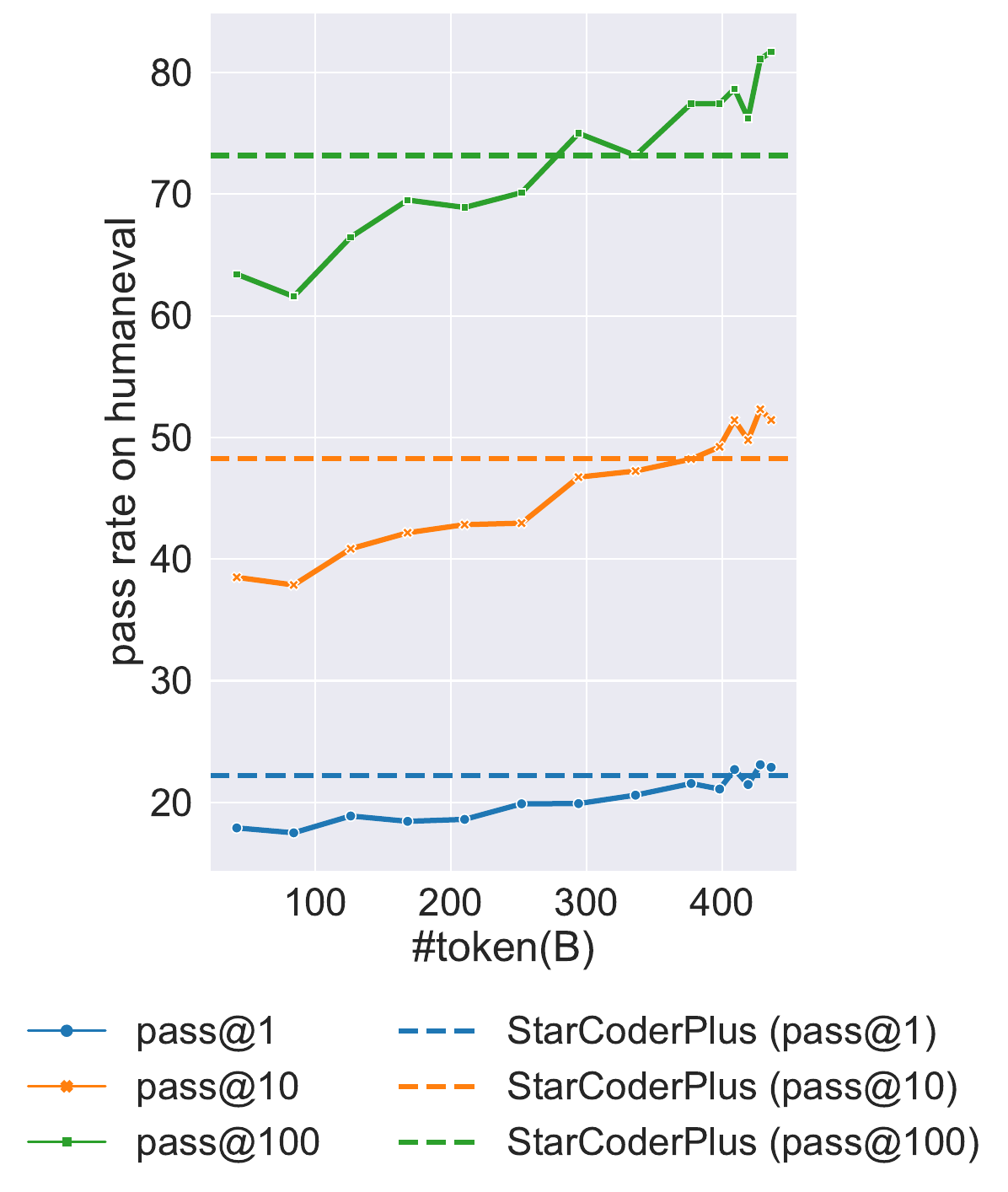}
        \vspace{0.42em}
        \caption{HumanEval}
        \label{fig:trend-he}
    \end{subfigure}
    \hfill
    \begin{subfigure}{0.3\textwidth}
        \centering
        \includegraphics[width=\textwidth]{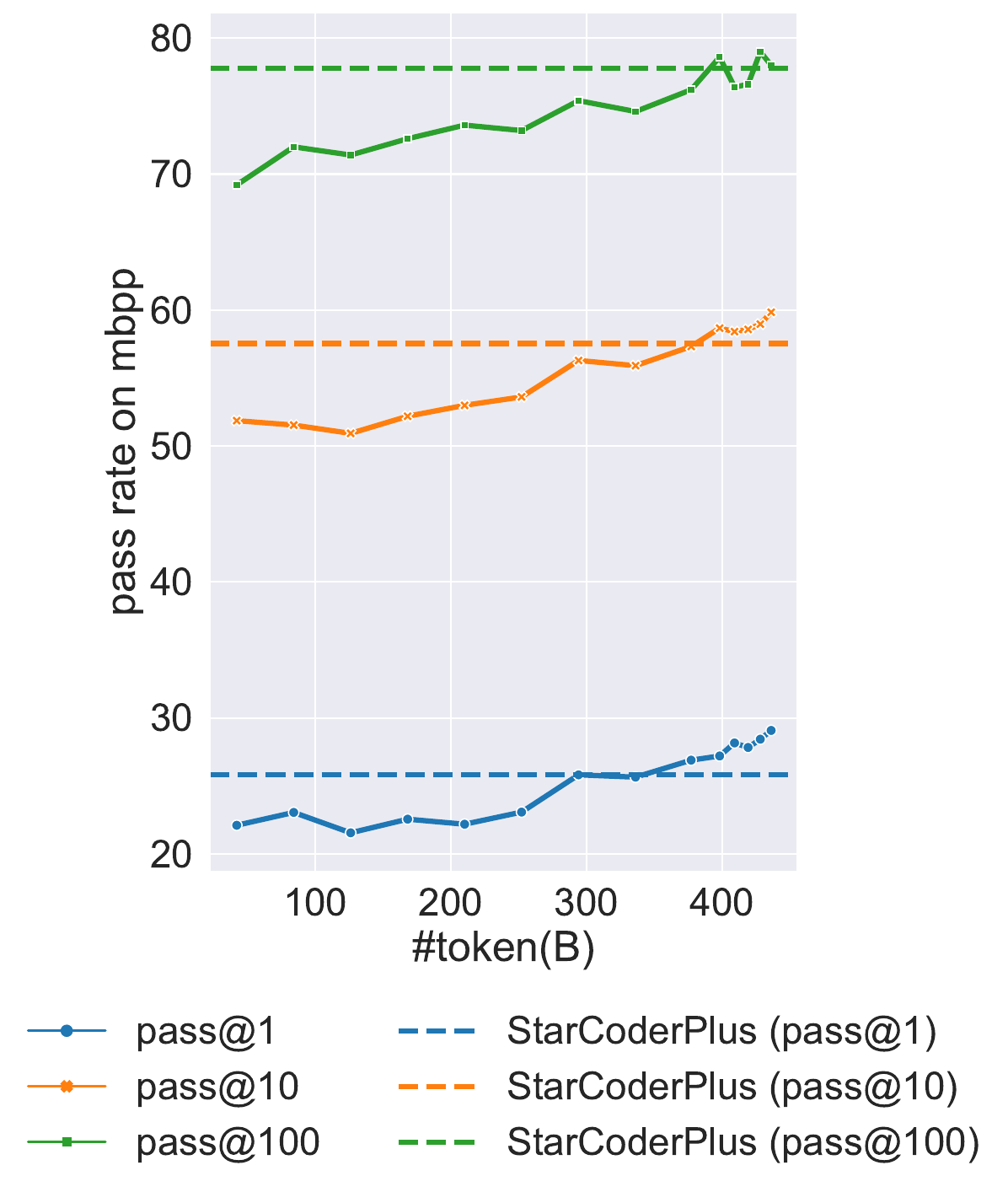}
        \vspace{0.42em}
        \caption{MBPP}
        \label{fig:trend-mbpp}
    \end{subfigure}
    \hfill
    \begin{subfigure}{0.33\textwidth}
        \centering
        \includegraphics[width=\textwidth]{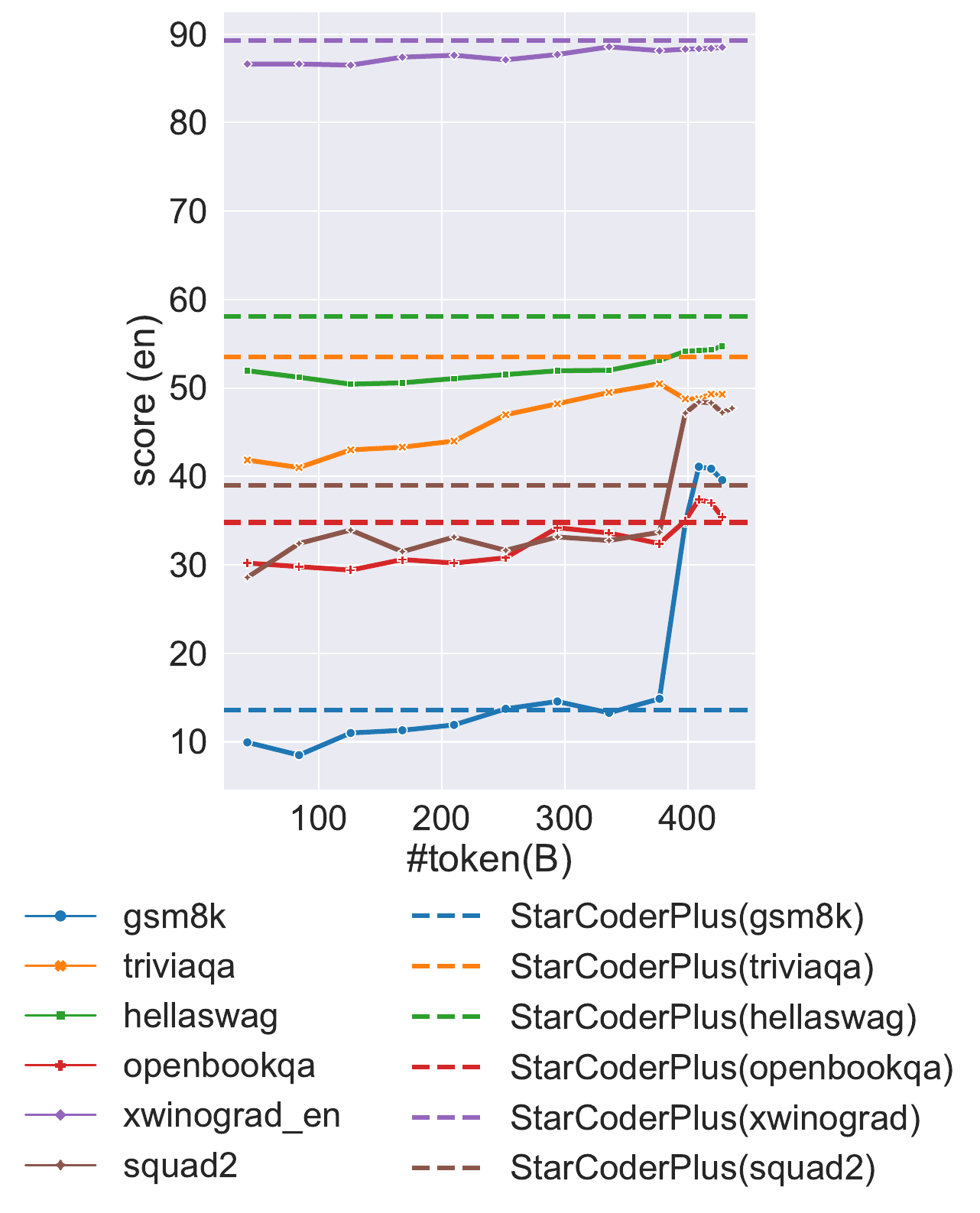}
        \caption{English}
        \label{fig:trend-en}
    \end{subfigure}
    \caption{Performance trends of models on HumanEval, MBPP, and English language tasks.}
    \label{fig:ana-code-en}
\end{figure*}

\begin{figure*}[t]
    \centering
    \begin{subfigure}{0.2\textwidth}
        \centering
        \includegraphics[width=\textwidth]{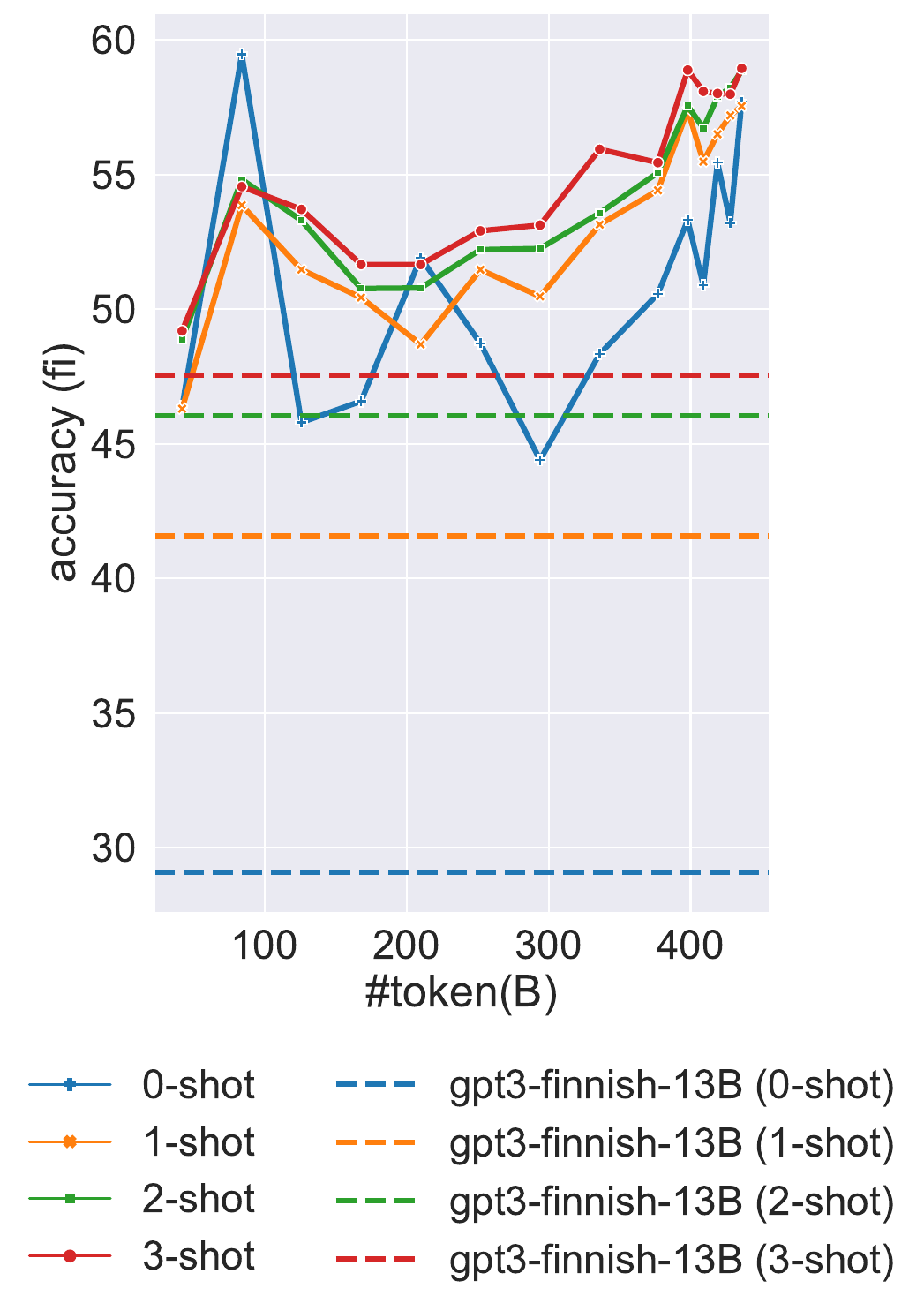}
        \caption{Finnish}
         \label{fig:trend-fi}
    \end{subfigure}
       \hfill
    \begin{subfigure}{0.24\textwidth}
        \centering
        \includegraphics[width=\textwidth]{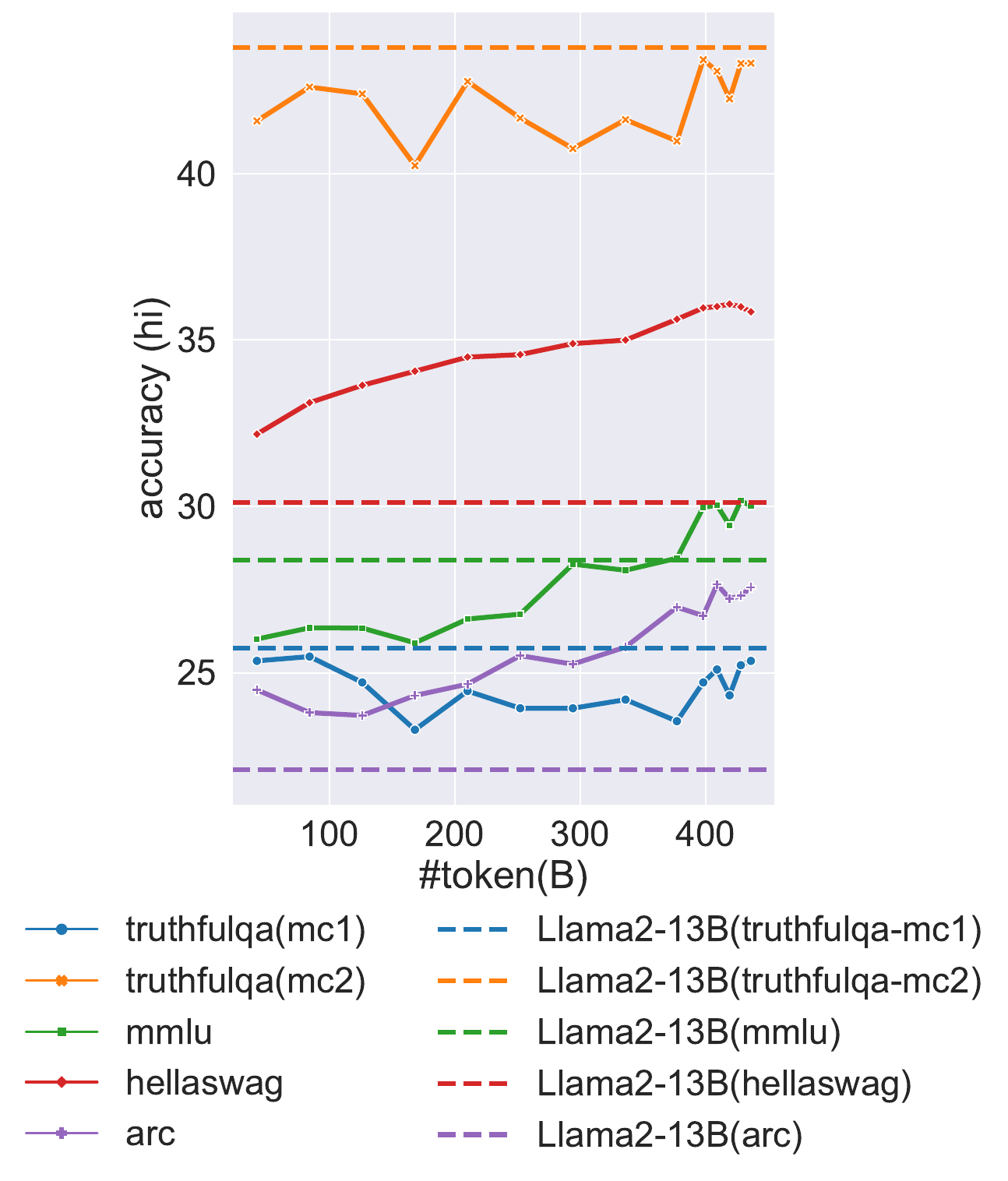}
        \caption{Hindi}
        \label{fig:trend-hi}
    \end{subfigure}
    \begin{subfigure}{0.24\textwidth}
        \centering
        \includegraphics[width=\textwidth]{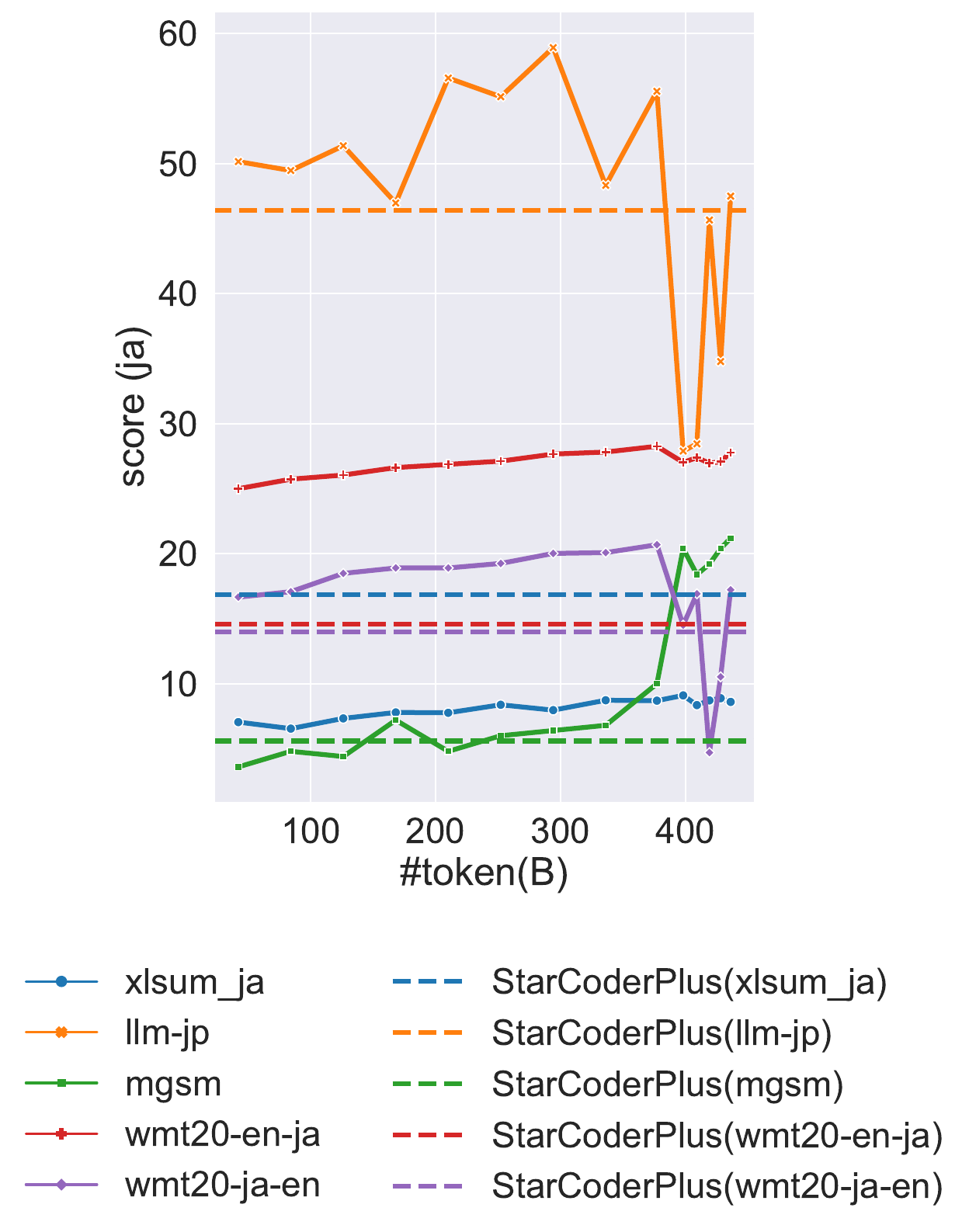}
        \caption{Japanese}
         \label{fig:trend-ja}
    \end{subfigure}
        \hfill
    \begin{subfigure}{0.24\textwidth}
        \centering
        \includegraphics[width=\textwidth]{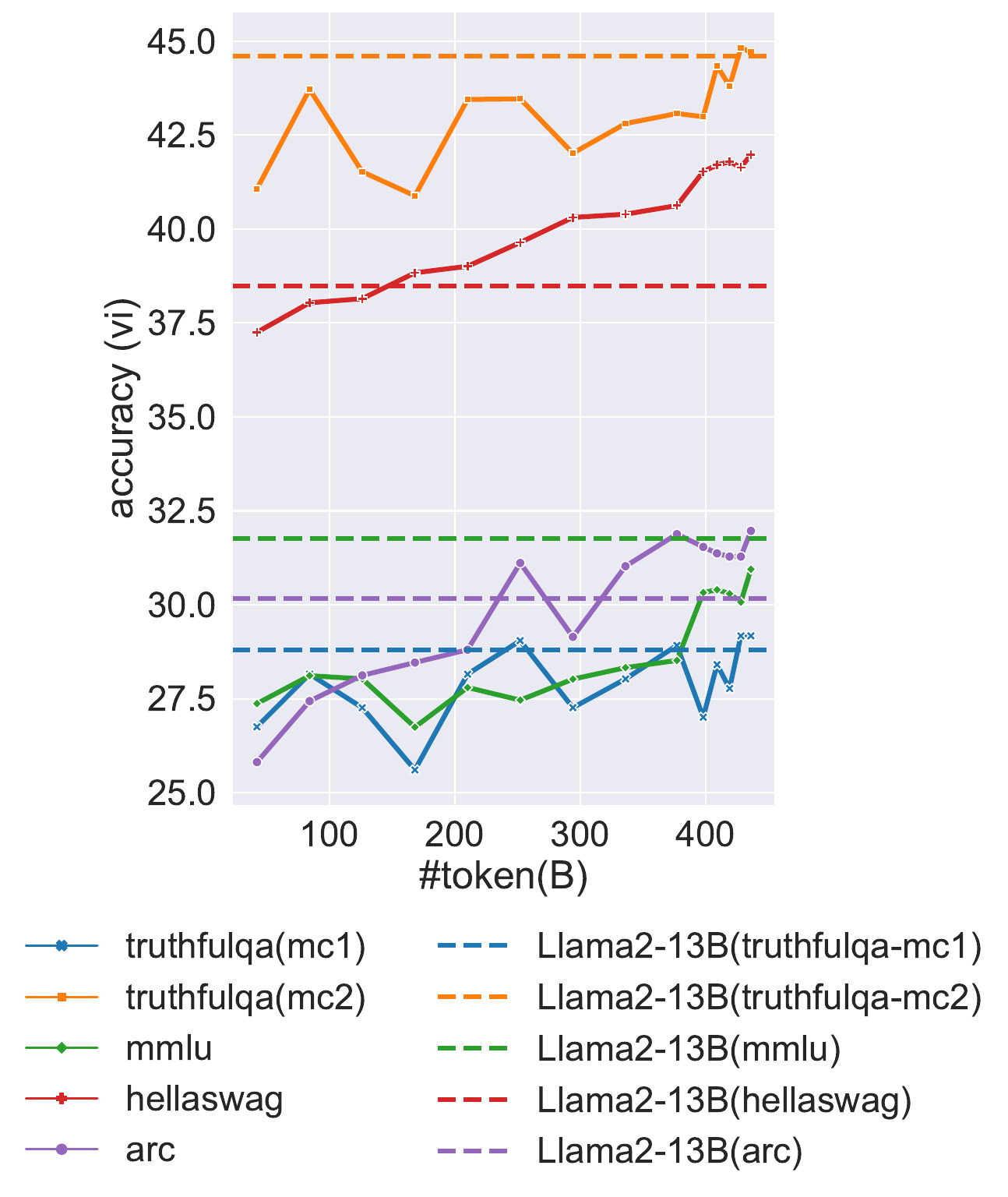}
        \caption{Vietnamese}
         \label{fig:trend-vi}
    \end{subfigure}
    \caption{Language-specific performance trends with increasing training tokens. Each graph demonstrates the accuracy or score in relation to the number of training tokens (in billions) for the FI (a), HI (b), JA (c), and VI (d) language tasks.}
    \label{fig:ana-lg}
\end{figure*}


\paragraph{Additional Safety Evaluations}\label{app:dangerous_qa}
Figure~\ref{fig:safety_dangerous} demonstrates our results on the DangerousQA dataset. Figure~\ref{fig:safety_redteam} shows the CARP values improving for our red-teamed \system. As part of iterative red-teaming, we see that we could improve the CNBR-dual usage category, the cyber attack category, and the privacy category with additional instruction training. 

\paragraph{Redteam Volunteers Protocol} Five of the authors volunteered to review and edit the generated responses from \system\ to create a subset of the Biden-Harris Redteam dataset, by editing for Biden-Harris concern violations and hateful, toxic, or bias output. One of the original volunteers and three other authors also provided CARP scores for \system\ responses to the Biden-Harris Redteam Testset shown in Figure~\ref{fig:safety_redteam}. Each volunteer is a machine learning professional over 18 years old and was informed of the risk of the sensitive subject matter of the responses.  Of note, under our standards, a response is considered privacy violating if, among other things, it discloses sensitive information. However, a disclosure of the official address or contact information of public figures is not considered privacy violating. 

\subsection{Performance Trends versus Training Token Compute}
\label{ap:ana}
Figure \ref{fig:ana-code-en} and \ref{fig:ana-lg} show on the relationship between the number of training tokens and the performance of the various models. This analysis aims to capture these trends for the code generation tasks such as HumanEval and MBPP, as well as for the English, Finnish, Hindi, Japanese, and Vietnamese language evaluations.

Starting with the HumanEval and MBPP evaluations (Figures \ref{fig:trend-he} and \ref{fig:trend-mbpp}), it is evident that the pass rates improve as the number of tokens increases. This suggests that the models are benefiting from more extensive training data, which likely includes a richer variety of programming challenges and solutions that enhance the model's problem-solving abilities. Notably, the Pass@100 rate for HumanEval shows a pronounced increase, indicating that, given enough attempts, the model has a high probability of generating a correct solution. This is consistent with the iterative nature of programming, where developers often refine their code through multiple iterations.

In the English language task (Figure \ref{fig:trend-en}), there is a marked variance in performance across different tasks as the number of tokens increases. The performance on GSM8K suddenly increases, which is attributed to the effect of the instruction tuning of our second training stage (CAT). Meanwhile, TriviaQA and Hellaswag tasks show steady improvements, indicating that these tasks may be benefiting more from the increased volume of training data.

The evaluations of the Finnish (FI) (Figure \ref{fig:trend-fi}), Hindi (HI) (Figure \ref{fig:trend-hi}), Japanese (JA) (Figure \ref{fig:trend-ja}), and Vietnamese (VI) (Figure \ref{fig:trend-vi}) languages reveal a similar trend of performance improvement with the increase in the number of tokens. However, there are some variances that might be attributed to the specific challenges each language presents, such as syntactic and semantic complexities. For instance, in the Finnish graph, the performance across different shot settings indicates that the model's ability to generalize from few examples improves with more data, which is a desirable trait in language models.

The evaluations for Japanese and Vietnamese exhibit an overall positive trajectory, albeit with intermittent fluctuations. These patterns suggest the potential for sustained incremental improvement through further continual pretraining on such datasets. However, due to computational
constraints, the extended pretraining is left for future work.

\end{document}